%% file: 0_main.tex
\documentclass[12pt]{elsarticle}

\usepackage{amssymb}
\usepackage{amsmath}
\usepackage{wrapfig}
\usepackage{listings}
\usepackage{todonotes}
\usepackage{listings}
\usepackage{multirow}
\newcommand{\hide}[1]{{}}
\usepackage{caption}
\usepackage{url}
\usepackage{rotating}
\usepackage{float}
\usepackage{afterpage}
\usepackage{array}
\usepackage{amsmath}
\usepackage{amssymb}
\usepackage{tcolorbox}
\usepackage{listings}
\usepackage{xcolor}
\usepackage{booktabs}

\definecolor{queryblue}{RGB}{240,248,255} % Light blue background
\definecolor{queryborder}{RGB}{70,130,180} % Steel blue border

\usepackage{courier}

\lstdefinestyle{sparqlstyle}{
  language=SQL,
  basicstyle=\ttfamily\small,
  keywordstyle=\color{blue}\bfseries,
  stringstyle=\color{red},
  commentstyle=\color{green!60!black},
  numbers=left,
  numberstyle=\tiny\color{gray},
  numbersep=5pt,
  frame=single,
  breaklines=true,
  breakatwhitespace=true,
  showstringspaces=false,
}

\definecolor{claudebackground}{RGB}{64,64,64} % Medium grey background
\definecolor{claudetext}{RGB}{229,229,229} % Light grey text
\definecolor{claudeborder}{RGB}{22,22,22} % Dark border

\lstdefinestyle{claudesparqlstyle}{
  language=SQL,
  basicstyle=\ttfamily\small\color{claudetext},
  keywordstyle=\color{claudetext}\bfseries,
  stringstyle=\color{claudetext},
  commentstyle=\color{claudetext},
  numbers=none,
  showstringspaces=false,
  breaklines=true,
  breakatwhitespace=true,
  frame=none,
}

\definecolor{grayishbackground}{RGB}{220,220,220}
\definecolor{basecolor}{RGB}{70,70,70}
\definecolor{factualcolor}{RGB}{180,0,0}
\definecolor{moralcolor}{RGB}{0,0,180}
\definecolor{symboliccolor}{RGB}{0,100,0}
\definecolor{impliedcolor}{RGB}{100,0,100}

\newcommand{\basetext}[1]{\textcolor{basecolor}{#1}}
\newcommand{\factualtext}[1]{\textcolor{factualcolor}{#1}}
\newcommand{\moraltext}[1]{\textcolor{moralcolor}{#1}}

\tcbset{
    basestyle/.style={
        colback=grayishbackground,
        coltext=basecolor,
        colframe=basecolor!30,
        boxrule=0.5pt,
        arc=3mm,
        left=3mm,
        right=3mm,
        top=2mm,
        bottom=2mm,
        fontupper=\ttfamily,
        fontlower=\ttfamily,
        title={\sffamily\bfseries\thetitle},  % Style the title here
    },
}

\journal{Information Processing \& Management}

\makeatletter
\def\ps@pprintTitle{%
 \let\@oddhead\@empty
 \let\@evenhead\@empty
 \def\@oddfoot{\reset@font\hfil\thepage\hfil}%
 \let\@evenfoot\@oddfoot
}
\makeatother
\begin{document}

\begin{frontmatter}



\title{Neurosymbolic Graph Enrichment for Grounded World Models}


\author[1]{Stefano De Giorgis} %% Author name
\author[1,2]{Aldo Gangemi}
\author[1]{Alessandro Russo}

\affiliation[1]{organization={Institute for Cognitive Sciences and Technologies - National Research Council (CNR-ISTC)},%Department and Organization
            %addressline={via Paolo Gaifami 18}, 
            %city={Catania},
            %postcode={95126}, 
            %state={CT},
            country={Italy}}

\affiliation[2]{organization={Department of Philosophy - University of Bologna},%Department and Organization
            %addressline={via Paolo Gaifami 18}, 
            %city={Catania},
            %postcode={95126}, 
            %state={CT},
            country={Italy}}

%% Abstract
\begin{abstract}
The development of artificial intelligence systems capable of understanding and reasoning about complex real-world scenarios is a significant challenge.
In this work we present a novel approach to enhance and exploit LLM reactive capability to address complex problems and interpret deeply contextual real-world meaning. We introduce a method and a tool for creating a multimodal, knowledge-augmented formal representation of meaning that combines the strengths of large language models with structured semantic representations. Our method begins with an image input, utilizing state-of-the-art large language models to generate a natural language description. This description is then transformed into an Abstract Meaning Representation (AMR) graph, which is formalized and enriched with logical design patterns, and layered semantics derived from linguistic and factual knowledge bases. The resulting graph is then fed back into the LLM to be extended with implicit knowledge activated by complex heuristic learning,
%. This comprehensive pipeline enables semantic enrichment across multiple dimensions, 
including semantic implicatures, moral values, embodied cognition, and metaphorical representations. By bridging the gap between unstructured language models and formal semantic structures, our method opens new avenues for tackling intricate problems in natural language understanding and reasoning.

\end{abstract}

% \begin{figure}
%     \centering
%     \includegraphics[width=\linewidth]{pipeline.png}
%     %\caption{Caption}
%     %\label{fig:enter-label}
% \end{figure}

% \begin{highlights}
% \item It is presented a novel neurosymbolic approach combining Large Language Models and knowledge graphs to create Extended Knowledge Graphs (XKGs) that capture implicit contextual knowledge across multiple semantic dimensions.
% \item The method introduces 11 knowledge heuristics, including presuppositions, conversational implicatures, and image schemas, to enrich formal representations of meaning derived from multimodal inputs like images and text.
% \item A comprehensive three-tiered evaluation framework is employed, encompassing logical validation, foundational ontology alignment, and human assessment, demonstrating the efficacy and potential of the approach for capturing nuanced semantic information.
% \end{highlights}

%% Keywords
\begin{keyword}
Neurosymbolic AI \sep Knowledge Representation \sep Knowledge Extraction \sep Large Language Models \sep Graph RAG \sep Hybrid Reasoning



\end{keyword}

\end{frontmatter}



\section{Introduction}
\label{sec:intro}

\input{1_intro}

\section{Related Work}
\label{sec:related}
\input{2_related}

\section{Methodology}
\label{sec:methodology}
\input{3_methodology}

\section{Experimental Evaluation}
\label{sec:evaluation}
\input{4_evaluation}

\section{Ongoing and Future Work}
\label{sec:ongoing}

\input{5_ongoing}

\section{Conclusions}
\label{sec:conclusions}

\input{6_conclusions}
\section*{Acknowledgements}
This work was supported by the Future Artificial Intelligence Research (FAIR)
project, code PE00000013 CUP 53C22003630006.

%\begin{thebibliography}{00}

%\bibliographystyle{plain} %elsart-num
%\bibliography{bib}
%% For numbered reference style

%\printbibliography

\end{document}

%% file: 1_intro.tex
The development of artificial intelligence systems capable of understanding and reasoning about complex real-world scenarios remains a significant challenge in computer science. This challenge is particularly evident when considering the distinct paradigms of generative AI and knowledge-based AI. Generative AIs, including Large Language Models (LLMs), function as signal processing machines: they learn activation patterns from input of any modality and provide symbolic output at inference time. In contrast, knowledge-based AIs operate as logical machines, extracting and designing symbolic representation patterns of the world with model-theoretical interpretations to ensure correct inference.

While both approaches have demonstrated remarkable capabilities, they often lack the comprehensive understanding necessary to build Grounded World Models (GWM) \cite{forrester1971counterintuitive,ha2018world}, i.e., models of the world as experienced or constructed by humans (and other organisms). 
GWMs are multivaried, encompassing physical, neurocognitive, social, and cultural dimensions. This multidimensional approach is intended to better mirror the complexity of human understanding.

This paper presents a novel neurosymbolic approach that aims to bridge this gap by leveraging LLMs' power of automated generation from latent knowledge, and the heuristic power of ontology-based knowledge graphs.
%to foster the development of truly Grounded World Models (GWMs).

At the core of our approach is the assumption that human-like understanding requires the integration of multiple layers of knowledge, ranging from sensorimotor experiences to abstract conceptual structures.
As argued by Lakoff and Johnson \cite{lakoff1999philosophy}, human cognition is grounded in embodied experiences, which give rise e.g. to image schemas and conceptual metaphors that shape our understanding of more abstract domains. An interesting consequence, which was not noticed until the appearance of LLMs \cite{nolfi2023unexpected}, is that abstract knowledge (as represented in natural language, logical and statistical models, knowledge bases, sensor data) can work as a supramodal system that bears correspondences to the physical, social, cognitive worlds.

Consequentially, we propose a framework for developing GWMs trained with, and capable of generating, knowledge graphs. Training includes either in-context learning or fine-tuning of pre-trained (multimodal) LLMs reinforced to generate multilayered knowledge, including highly contextual implicit knowledge. Implicit knowledge includes e.g. presuppositions, conversational implicatures, factual impacts, image schemas, metonymic and symbolic coercions, event sequences, causal relations, and moral value-driven reasoning.

A key innovation in our approach is the use of LLMs not as expert systems, but as reactive engines to extract implicit contextual knowledge. 
%As Kautz \cite{kautz2022third} notes, %the current ``third AI summer'' is characterized by the emergence of large neural models that capture broad patterns of world knowledge in a distributed, subsymbolic form. 
Besides using heuristic rules and  curated knowledge bases, we leverage the reactive power embedded in LLMs to extract and formalize general knowledge across multiple semantic dimensions. %This allows us to address longstanding challenges in AI, such as the ``Frame problem'', by tapping into the vast reservoir of everyday knowledge and reasoning patterns learned by LLMs from diverse textual corpora.

To operationalize this vision, we introduce a novel method to create knowledge-augmented formal representations of multimodal meaning. For example, starting from an image input, we reuse state-of-the-art LLMs to generate a natural language description. This description is then transformed into Abstract Meaning Representation (AMR) graphs \cite{bevilacqua-navigli-2020-breaking} which are then formalised as an ontology-based knowledge graph \cite{gangemi2017semantic}, and aligned to public knowledge bases, with the Text2AMR2FRED tool \cite{meloni2017amr2fred,gangemi2017amr2fred,gangemi2023taf}.
%generating a well formed RDF graph, retrieving entities from well known semantic web resources. 
The resulting graph is iteratively extended with implicit knowledge activated by heuristics for presuppositions, coercions, impact, etc.
%corresponding to , creating a rich, multi-faceted representation of knowledge.

%Importantly, o
Our approach implements a feedback loop that leverages multimodal, multilayered GWMs. This iterative process enables continuous refinement and expansion of the knowledge graph, enhancing the system's ability to adapt to new contexts and improve its understanding over time.

%Crucially, our 
This neurosymbolic approach makes the process of knowledge base enrichment  more agile and scalable compared to traditional methods. By combining the flexibility and generative power of LLMs with the structure and inferential capabilities of symbolic knowledge representations, we enable rapid expansion and refinement of knowledge graphs across diverse domains. 

This hybrid architecture can be classified as a Type 2-3 neurosymbolic system in Kautz' taxonomy \cite{kautz2022third}.
%is ready for the dynamic integration of new knowledge and adaptation to novel contexts.

The potential applications of this approach are far-reaching, spanning natural language understanding, visual reasoning, and complex problem-solving tasks. By providing a framework for grounding abstract language in embodied and commonsense knowledge, our method opens new avenues for developing AI systems capable of human-like reasoning and contextual understanding. %Furthermore, the modular nature of our method allows for continuous refinement and expansion of the knowledge layers considered, paving the way for increasingly sophisticated and nuanced world models.

In the following sections, we detail the technical architecture of our implemented system and a short introduction to a public demo, and present experimental results demonstrating the efficacy of our approach across multiple knowledge domains.

The paper is organised as follows: Section \ref{sec:related} provides state-of-the-art positioning, Section \ref{sec:methodology} details the methodology, Section \ref{sec:evaluation} describes the three-tier evaluation of the method, Section \ref{sec:ongoing} illustrates ongoing and future works, and finally Section \ref{sec:conclusions} wraps up the paper.

%% file: 2_related.tex
Our approach is positioned in the quickly growing field of Large Language Models and Knowledge Graphs hybrid neurosymbolic methods.
%interactive conjoint approaches.
%Here we propose some useful pointers to state of the art literature, while we describe our work's modular architecture in Section \ref{sec:methodology}.
The synergy between neural and symbolic approaches has been extensively explored in multiple surveys \cite{yu2023survey,belle2020symbolic,besold2021neural,garcez2023neurosymbolic,sarker2021neuro}, some of them focusing explicitly on graph neural networks and symbolic components \cite{lamb2020graph}, reasoning over graph structures \cite{zhang2021neural}, dynamic knowledge graphs \cite{alam2024neurosymbolic}, and natural language inference \cite{yao2018learning}, reflecting the growing interest in hybrid methodologies.

\begin{wrapfigure}{l}{0.7\linewidth}
    \centering
    \includegraphics[width=\linewidth]{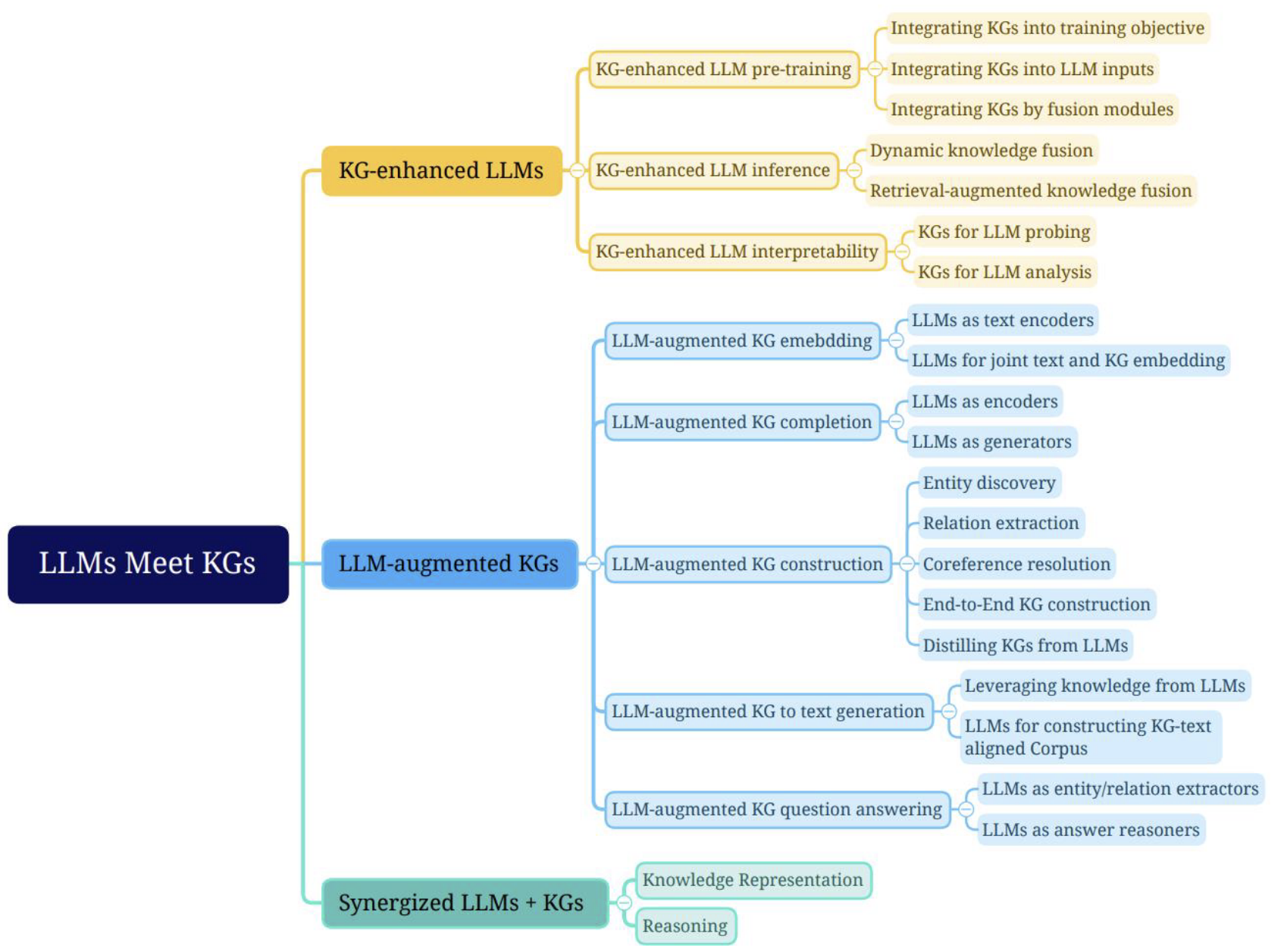}
    \caption{Roadmap for LLMs and KGs interactions, taken from \cite{pan2024unifying}.}
    \label{fig:roadmap}
\end{wrapfigure}

While an exhaustive survey 
%examination of this domain 
lies beyond the scope of this paper,
%our current work, 
%it is pertinent to contextualise our approach 
we can position our method within the broader landscape of recent advancements: (i) 
%the positioning of our work 
within the landscape of interactions between LLMs and KGs in joint methods
%conjoint approaches
, and (ii) in relation to popular techniques \\ such as Graph RAG \\ (Retrieval-Augmented Generation) that have recently garnered significant attention. 

Regarding joint LLMs and KGs methods, according to \cite{pan2024unifying}, as shown in Figure \ref{fig:roadmap}, our approach would be both ``LLM augmented KG'' and ``Synergized LLMs + KGs'', and in particular ``LLMs - augmented KGs construction'' focusing on ``End-to-end construction'' and ``Distilling KGs from LLMs'', as well as ``LLMs-augmented KGs to text generation'' focusing on ``Leveraging knowledge from LLMs'', and partially ``LLM-augmented KG for question answering'' adopting ``LLMs as entity/relations extractors''.

\noindent Considering Kautz' taxonomy of neurosymbolic systems \cite{kautz2022third}, as well as Hamilton contextualization in the domain of natural language understanding \cite{hamilton2022neuro}, our pipeline qualifies as Type 2 (``Symbolic[Neuro] (Nested)'') - Type 3 (``Neuro; Symbolic (Cooperative)'') architecture.
Type 2 is defined as nested architecture, where a symbolic reasoning system is the primary system with neural components driving certain internal decisions, while Type 3 covers cases in which a neural network focuses on one task, e.g. entity linking, word sense disambiguation, etc., and interacts via input/output with a symbolic reasoner specializing in a complementary task, in our case, building semantic tree dependencies in RDF format while aligning entities to Framester \cite{gangemi2016framester} hub of ontologies, maintaining correct OWL2 syntax.

%To what concerns the connection with
Considering Graph RAG approaches, in recent years the integration of LLMs and knowledge graphs has gained significant traction \cite{zhu2024llms}, particularly using models for knowledge graph completion \cite{yao2019kg}, and through the development of Retrieval-Augmented Generation (RAG) and Graph RAG techniques. Different models (GPT, Llama, Claude, Gemini, etc.) have shown different specific capabilities but similar structural limitations \cite{meyer2023llm}.
While traditional RAG focuses on retrieving precise information from unstructured or vector-compressed resources, Graph RAG has demonstrated remarkable efficacy and compliance with graph-structured data \cite{edge2024local}. The Graph RAG approach extends beyond mere text vectorisation, encompassing a two-fold process: entity recognition and relation extraction, culminating in the generation of semantic triples.

Classic applications of Graph RAG have been in question-answering \cite{sen2023knowledge}, starting from textual information transposed to knowledge graph embeddings \cite{lu2020utilizing}, or summarization systems designed to retrieve specific information from established knowledge bases \cite{edge2024local}.
However, certain layers of meaning remain challenging to capture and formalize, including moral reasoning, pragmatic implicatures, and tacit knowledge derived from real-world experiences. 

The topic of knowledge graph generation from text is of some relevance in several overlapping communities, such as the Semantic Web one \cite{tiwari2022preface}, as well as the broader knowledge-graph-oriented one\footnote{\url{https://www.linkedin.com/posts/tonyseale_the-microsoft-graphrag-library-has-recently-}\\ \url{activity-7230121112158830593-m8At?utm_source=share&utm_medium=member_desktop}} \cite{edge2024local}.

For this reason, our research expands beyond Graph RAG. 
We leverage LLMs as latent reactors that can provide approximate, implicit, commonsense knowledge when activated with appropriate heuristics, rather than as tools for compressed information retrieval.
Our approach, partially positioned as Knowledge Augmented Generation (KAG)\footnote{For terminological reference, cf.: \url{https://aidanhogan.com/talks/2024-09-04-wuwien-invited-talk.pdf}}, enables us to extract and formalise implicit information that humans intuitively infer, especially in visual comprehension tasks. 

By doing so, our work addresses a longstanding challenge in artificial intelligence – the Frame problem\footnote{\url{https://plato.stanford.edu/entries/frame-problem/}} – which concerns the ability of AI systems to contextualise information and make human-like inferences, massively reducing the combinatorial space of choices.
Several past approaches, such as the work of Brachman \cite{brachman1977s}, Minsky \cite{minsky1974framework}, and Fillmore \cite{fillmore2006frame} attempted to tackle this issue by using structured representation.
Being able to faithfully capture the nature of a domain without enumerating all the implicit knowledge that it assumes when discussing or performing practices is indeed a cognitive and computationally hard task.

Concerning technical components, we include here some knowledge extraction tools/methods that have paved the way to what we present in this article.

FRED\footnote{Available at \url{http://wit.istc.cnr.it/stlab-tools/fred/demo/}} \cite{gangemi2017semantic} is a comprehensive tool for formal knowledge graph generation from natural language (using the OWL2 description logic \cite{grau2008owl}), including additional features such as entity recognition and entity linking, frame extraction, semantic role labeling, and word sense disambiguation.
FRED uses Categorical Grammar \cite{lambek1988categorial} and Boxer \cite{bos2008wide} to create a formal graph, which is then aligned to WordNet, FrameNet, DBpedia, and other public resources.

Abstract Meaning Representation (AMR) has been proposed long ago in computational linguistics as a solution for natural language representation that can be pragmatically extracted from text. Recently, an efficient parser \cite{bevilacqua-navigli-2020-breaking} based on transformer's end-to-end technology has opened new ways to make it scalable.
AMR has a graph-based structure that informally encodes complex relationships and dependencies between concepts, facilitating a more nuanced representation of linguistic meaning. Compared to predecessors like Categorical Grammars, AMR abstracts away from surface-level syntactic variations, simplifying the application of logical and ontological patterns when creating knowledge graphs from text.
%allows for a standardized semantic representation across different phrasings of the same underlying idea, making it particularly valuable for our implicit knowledge extraction and commonsense reasoning tasks.

AMR2FRED \cite{gangemi2017amr2fred}, a deterministic, rule-based system, which applies the formal design patterns of FRED's to AMR, and provides alignments reusing Framester \cite{gangemi2016framester} as main hub of public resources.
The Framester ontology hub \cite{gangemi2016framester,gangemi2020closing} provides a formal semantics to semantic frames \cite{fillmore2006frame} in a curated linked data version of multiple linguistic resources (e.g. FrameNet \cite{baker1998berkeley}, WordNet \cite{miller1998wordnet}, VerbNet \cite{schuler2005verbnet}, PropBank \cite{kingsbury2002treebank}, a cognitive layer including MetaNet\cite{gangemi2018afaoocm} and ImageSchemaNet \cite{de2022imageschemanet}, BabelNet \cite{navigli2010babelnet}, factual knowledge bases (e.g. DBpedia \cite{auer2007dbpedia}, YAGO \cite{suchanek2007yago}, etc.), and ontology schemas (e.g. DOLCE-Zero \cite{gangemi2003sweetening}), with formal links between them, resulting in a strongly connected RDF/OWL knowledge graph.

Considering LLMs, the modular method we present here is potentially agnostic to the LLMs used. Nevertheless, the current version of the tool implementing the method, available for testing online\footnote{\url{https://arco.istc.cnr.it/itaf/}}, uses GPT-4o and Claude Sonnet 3.5 APIs. The results, shown in Section \ref{sec:evaluation}, are therefore based on these two top-scoring models.

%% file: 3_methodology.tex
\hide{
\begin{itemize}
    \item declarative textual description of the image from GPT
    \item re-inject the textual description in LLM (Claude) looking for the following semantic layers (add short definitions for each) - add table - add image
    \item explain why these
    \item one example of more ad hoc personalized feature to be extracted
    \item testing of the methodology (to be defined clearly)
\end{itemize}
}

\hide{
Problematic points:
- although it is pretty straightforward to generate kg from text, then that formalised knowledge should be interoperable, clear in its semantics, sharable, accessible. It has to be overcome the paradox of linguistic solipsism (Wittgenstein, Derrida).
In another words: th produced kg has to be aligned with existing ontologies, assuming the ontology is well modeled and reuses e.g. existing resources, Ontology Design Patterns, etc.
In our case we use DOLCE 0 to overcome these issues, and we do so via AMR2FRED tool.
}

In this section we describe our approach, detailing (i) the technical structure, and (ii) the knowledge layers we have chosen to include in our heuristics. It is important to note that the selection of the heuristics is not exhaustive and can be incrementally expanded to incorporate additional knowledge areas as needed, as envisioned in Section~\ref{sec:ongoing}.

\begin{figure}
    \centering
    \includegraphics[width=\linewidth]{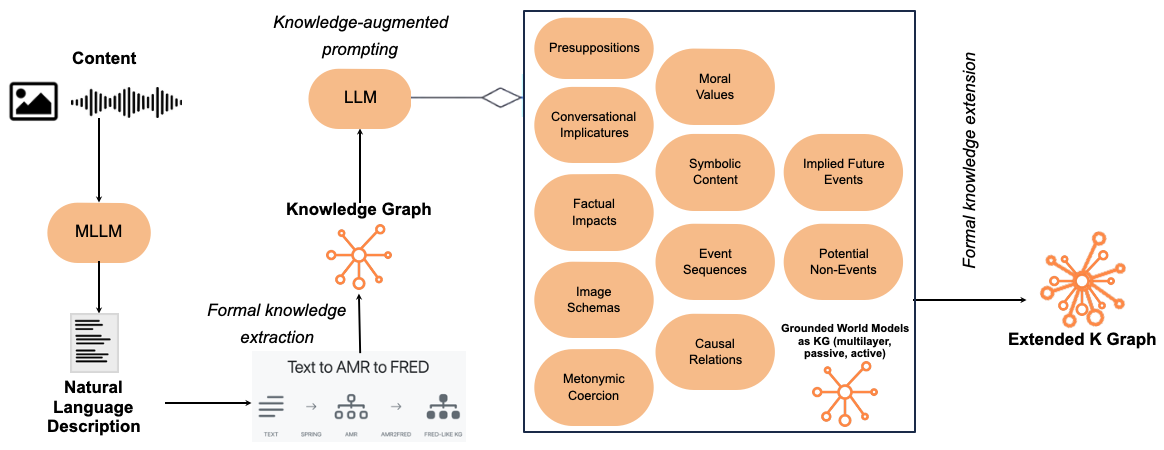}
    \caption{Hybrid knowledge enrichment pipeline.}
    \label{fig:pipeline}
\end{figure}

\paragraph{\textbf{XKG Generation Pipeline}}
Our modular pipeline combines natural language processing, knowledge representation, and large language models (LLMs) to extract and enrich semantic information from textual or visual inputs. The whole process is shown in Figure~\ref{fig:pipeline}, and begins on the top left, with either user-provided text or an LLM-generated description of an input image, using a carefully crafted prompt.

Considering the case in which we provide a picture as starting input, following Figure~\ref{fig:pipeline}, this original content is passed to a Multimodal LLM (MLLM), prompted to provide a description of the picture in natural language.

In our current implementation, after several tests, we retrieved that GPT-4o is the best model at providing a textual description when given as input an image.
Therefore, starting from an image, we rely on GPT-4o to get as output a textual description.
In Section~\ref{sec:evaluation} we provide an example of input and output; furthermore, all the prompts are provided as additional material on the GitHub repository\footnote{\url{https://github.com/StenDoipanni/XKG/tree/main/prompt}}.

This textual representation is passed to the Text2AMR2FRED (TAF) tool~\cite{gangemi2023taf}, which transforms the input text into an Abstract Meaning Representation (AMR) graph using the SPRING parser~\cite{bevilacqua2021one}. As part of this step, entity linking is performed from the nodes of the AMR graph to Wikipedia entries using BLINK~\cite{wu2019scalable}. Due to the opaque nature of the Text-to-AMR conversion, which doesn't explicitly map text segments to graph elements, entity linking is performed heuristically: BLINK identifies Wikipedia entity mentions in the text, and if AMR nodes are labeled with matching text segments, they are linked to the corresponding Wikidata entities.

The AMR graph is subsequently converted to an OWL-RDF knowledge graph relying on AMR2FRED~\cite{gangemi2017amr2fred,meloni2017amr2fred} using FRED heuristics~\cite{gangemi2017semantic} and motifs \cite{gangemi2016kbs}.
The enrichment of this RDF graph through word sense disambiguation (WSD) is done using the eWiSeR tool~\cite{bevilacqua-navigli-2020-breaking}. Similar to the entity linking process, WSD is applied to the original text, linking text segments to WordNet synsets (sets of contextual synonyms). In essence, when RDF graph entities have names (roughly corresponding to the final part of their URIs) that match disambiguated text segments, an \texttt{owl:equivalentClass} triple linking the entity to the WordNet synset is added to the graph. Leveraging Framester~\cite{gangemi2016framester}, we augment the graph with additional \texttt{rdfs:subClassOf} triples for the disambiguated entities, connecting them to WordNet supersenses and DOLCE types, derived from querying Framester with the identified WordNet synsets.
This RDF graph, the result of the text-to-AMR-to-FRED (TAF) pipeline, serves as our ``Base Graph'', shown in Figure \ref{fig:pipeline} as ``Knowledge graph'' in the middle of the pipeline.

From here, we re-inject this Base Graph to LLM for knowledge enrichment.
For this step, after experimenting with a gold standard of formal knowledge graphs extracted from text \cite{gangemi2016kbs}, we found that Claude 3.5 Sonnet is the best model for the purpose.
The ``Knowledge Augmented prompting'' step of the pipeline calls Claude 3.5 Sonnet API, and takes as input a specific base prompt, including the Base Graph expressed in Turtle syntax. This base prompt is then refined in a specific prompt for each heuristic, shown as salmon-colored boxes on the right of Figure~\ref{fig:pipeline}.

For each predefined enrichment heuristic, we prompt Claude to generate additional triples, expanding the implicit knowledge captured in the graph, anchoring the newly introduced triples to nodes of the Base Graph. Each heuristic produces a separate graph that combines the Base Graph with the heuristic's specific additional triples. Finally, we construct a comprehensive graph that merges the base graph with all the additional triples generated across all heuristics, resulting in a richer, multi-faceted semantic representation of the initial image or text input.

\subsection{Knowledge Heuristics}

We describe here the 11 heuristics currently implemented in the implemented tool. 
%Heuristics are used to enrich the Base Graph in our multimodal Knowledge Augmented Graph (KAG) model.
These heuristics are chosen according to an analysis of the essential elements that constitute daily human understanding and sense-making activity in building and using Grounded World Models. The heuristics list includes: \textit{Presuppositions}, based on previous background knowledge; \textit{Conversational Implicatures}, which often contributes in making sense of incomplete information in linguistic exchanges; \textit{Factual Impact}, which grounds linguistic entities to factual knowledge; \textit{Image Schemas}, basic building blocks of cognition which grounds our way of conceiving the world in our sensori-motor bodily perception (grounding e.g. cognitive metaphors and several other entities); \textit{Metonymic Coercions}, which allows understand propositions whose truth value would be zero, but differ from metaphorical speech grounding the relation between entities on the parthood-whole relation; \textit{Moral Value Driven Coercion}, applied everyday in appraisal and moral evaluative processes, values nudge our daily behavior; \textit{Symbolic Coercion}, in Peirce terminology \cite{peirce1902logic}, used to anchor meaning to various entities of the world; \textit{Event Sequences}, determinant in our plan-making capability and ability to design plausible scenarios and outcomes; \textit{Causal Relations}, establishing relations of cause-effect between processes and events, to avoid either having only (i) temporal sequences and (ii) statistical correlation; \textit{Implied Future Events}, a specification of Event Sequences, for temporal projection in the future; and \textit{Implied Non-Events}, an infinite set of events, but, referring to the Frame problem, focusing on those more closely related to a specific Event Sequence.

\paragraph{\textbf{Presuppositions}}
Presuppositions\footnote{\url{https://plato.stanford.edu/entries/presupposition/}} are implicit assumptions necessary for statements to be meaningful \cite{strawson1950referring,frege1892sense,karttunen2016presupposition}. They play a crucial role in human cognition, enabling efficient communication through shared background knowledge. In natural language, presuppositions help infer unstated information and enhance contextual understanding. By formalizing and integrating these implicit meanings into knowledge structures, the approach captures nuanced understandings often taken for granted in human communication. For example, the statement ``The athlete won the gold medal'' presupposes that there was a competition, possibly an Olympic one, illustrating how presuppositions convey information beyond the explicit content of a sentence.

\paragraph{\textbf{Conversational Implicatures}}
\hide{mention Pragmatics Linguistics}
Conversational implicatures\footnote{\url{https://plato.stanford.edu/entries/implicature/}} are implied meanings that arise from the context of a conversation rather than literal interpretation. Defined in Grice's pragmatics \cite{grice1975logic}, they are essential to human communication, allowing speakers to convey more information than explicitly stated \cite{levinson2000presumptive}. By formalizing and integrating these implicatures into knowledge structures, the approach captures nuanced, context-dependent interpretations that humans naturally derive from conversations. This enhances the system's ability to understand and reason about complex linguistic phenomena. For example, if someone asks ``Is there a gas station nearby?'' and receives the answer ``There's one around the corner,'' the implicature here is that the gas station should be open and operational, even though this isn't explicitly stated.

\paragraph{\textbf{Factual Impact}}
Factual impact refers to the physical, social, and cognitive consequences of events on participants, including expected emotions, sensations, and changes in mental states \cite{kahneman1991article}. This concept is crucial for understanding the full implications of human interactions and narratives. By formalizing these impacts, the approach creates a comprehensive representation of how events affect individuals on multiple levels. This enriched representation allows for inference and reasoning about e.g. the emotional and cognitive states of participants, adding human-like comprehension to the analysis of social interactions. For example, in the event of ``winning a competition,'' the factual impact might include physical sensations (adrenaline rush), emotions (joy, pride), and cognitive changes (increased confidence, future goal-setting).

\paragraph{\textbf{Image Schemas}}
Image schemas are fundamental cognitive structures that help humans organize and interpret their experiences of the world \cite{Johnson87,lakoff1999philosophy}. These gestaltic schemas, such as container, path, balance, and force, underlie our perceptual understanding of sentences and situations \cite{besold2017narrative}. By integrating these schemas into knowledge representation, the approach captures a crucial aspect of human cognition and spatial reasoning \cite{hedblom2020image}. This allows for inference and reasoning about underlying spatial and conceptual structures that humans instinctively understand, also via cognitive metaphors \cite{lakoff1980metaphors}, enhancing the system's ability to represent complex narratives and interactions. For example, the container schema helps us understand phrases like ``in the competition'' or ``out of the box,'' while the path schema underlies our comprehension of sentences describing movement or progress.

\paragraph{\textbf{Metonymic Coercion}}
Metonymic coercion is a linguistic phenomenon where a word's typical or literal sense is overridden by a specific, related sense within its extended semantics. This process is fundamental to human language comprehension, enabling more efficient and nuanced communication. By formalizing metonymic coercions in knowledge representation \cite{maudslay2024chainnet}, the approach captures implicit semantic shifts that occur in everyday language. This allows the system to infer and reason about intended meanings behind metonymic expressions, bridging the gap between literal interpretations and context-dependent understandings. For example, in the sentence ``The White House announced new policies,'' ``White House'' undergoes metonymic coercion to represent the U.S. government or administration, rather than the physical building.

\paragraph{\textbf{Moral-Value-Driven Coercion}}
Moral-value-driven coercion\footnote{\url{https://plato.stanford.edu/entries/value-theory/}} is a linguistic phenomenon where the literal meaning of an action or statement is overridden by a morally or socially charged interpretation. This process reflects the underlying value systems \cite{graham2013moral,schwartz_extending_2001} that guide human behavior and decision-making \cite{rozin_cad_1999}. By formalizing these coercions in knowledge representation, the approach captures implicit moral and social dynamics at play in human interactions. This allows the system to infer and reason about underlying ethical considerations and social norms that shape human behavior, adding depth to the understanding of complex narratives and interpersonal dynamics. For example, the statement ``She always keeps her promises'' might be coerced from a literal description of behavior to a moral judgment about the person's integrity and trustworthiness within a social context.

\paragraph{\textbf{Symbolic Coercions}}
Symbolic coercions involve the transformation of literal meanings into symbolic interpretations\footnote{\url{https://plato.stanford.edu/entries/peirce-semiotics/}}. This process is fundamental to human cognition and communication \cite{peirce1902logic}, allowing for the anchoring of complex ideas to familiar concepts, objects, and imagery. By formalizing these coercions in knowledge representation, the approach captures implicit symbolic meanings that permeate human language and thought. This enables the system to recognize and reason about the underlying symbolic significance of actions and concepts, particularly in domains where abstract ideas are often expressed through concrete objective correlative \cite{eliot1920hamlet}. For example, in sport events commentaries, the phrase ``the cangaroos won the match'' undergoes symbolic coercion from a literal description of marsupial mammals to representing Australia (including potential biases and stereotypes).

\paragraph{\textbf{Event Sequences}}
Event sequences capture the chronological relationship between events mentioned in a text, providing crucial information about the order of actions or occurrences. By formalizing these sequences, the approach creates a rich representation of implicit chronological information in narratives. This enables the system to infer and reason about the temporal flow of events, even when not explicitly stated. Such temporal reasoning is essential for understanding causality, narrative progression, and the logical flow of actions and reactions in complex scenarios. For example, in the sentence ``After the flight, the athletes have two days before the competition,'' the system would recognize the sequence: $flight \rightarrow two\_days \rightarrow competition$, allowing for deeper comprehension of the narrative's timeline and potential causal relationships between events.

\paragraph{\textbf{Causal Relations}}
Causal relations capture the cause-and-effect relationships between events or states mentioned in a text. By formalizing these relationships the approach creates a rich representation of implicit causal information. This enables the system to infer and reason about underlying causes and effects within complex scenarios, even when not explicitly stated. Understanding causal relations is essential for comprehending motivations, predicting outcomes, and constructing coherent mental models of situations. For example, in the sentence ``The heavy rain forced the match to be postponed,'' the system would recognize the causal chain: $heavy\_rain \rightarrow match\_postponing$, allowing for deeper analysis of the event's progression and potential consequences.

\paragraph{\textbf{Implied Future Events}}
Implied future events involve inferring likely outcomes or consequences based on given information. This concept captures the human ability to anticipate future scenarios and make predictions based on current contexts and interactions. By formalizing these implied future events, we are able to reason about probable consequences of current actions and statements, even when not explicitly mentioned in the text. Such predictive reasoning is essential for understanding long-term significance of interactions. For example, in the sentence ``The committee announced more strict regulations,'' the system might infer potential future events such as more severe checks, increase in bureaucratic practices, increase in controversy, etc.

\paragraph{\textbf{Implied Potential Non-events}}
Implied potential non-events are events that could have occurred but are prevented or made unlikely due to other circumstances or decisions mentioned in the text. This concept captures the understanding of alternative scenarios and the implications of characters' choices. Although this is an infinite set of entities, namely all the possible alternative scenarios that will never take place given a starting situation, we focus on those implicit alternatives that humans naturally consider when interpreting narratives. This enables the system to reason about not just what happens, but also what could have happened under different circumstances. Such counterfactual reasoning is essential for understanding motivations, the significance of decisions, and the broader implications of narrative events. For example, in the sentence ``She decided not to compete,'' the system would recognize the implied potential non-event of participating to the competition, allowing for analysis of the decision's consequences and alternative outcomes.

\vspace{5mm}

\noindent The graphs produced from all these heuristics together form the set of Extended Knowledge Graphs (XKGs).

%% file: 4_evaluation.tex
In this section we evaluate the output generated from the full pipeline described in Section \ref{sec:methodology}: we provide an example of knowledge extension starting from an image to the complete XKGs extension with all 11 heuristics. 

\begin{wrapfigure}{r}{0.4\linewidth}
    \centering
    \includegraphics[width=\linewidth]{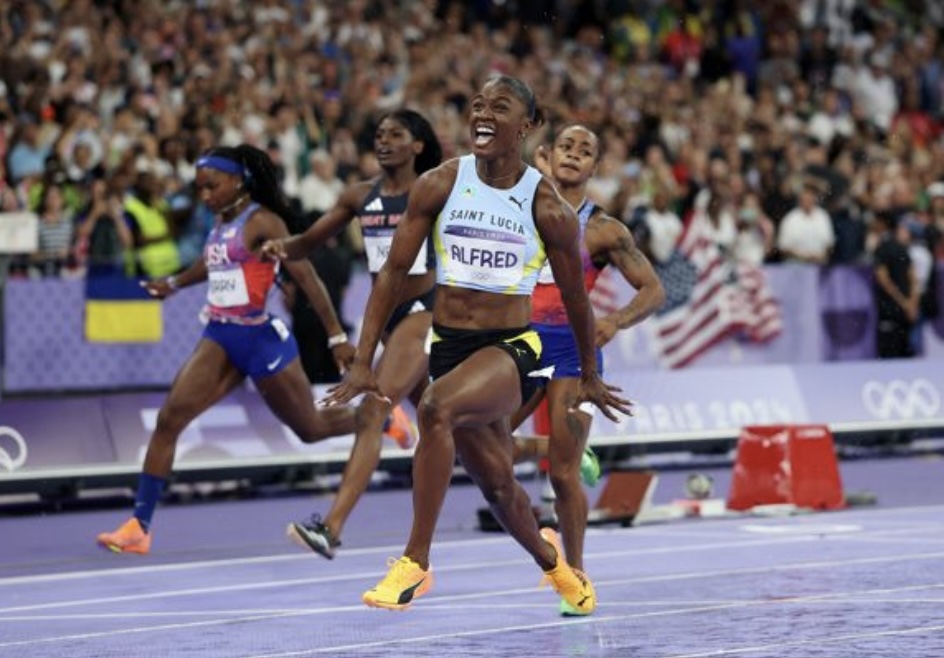}
    \caption{Gold medal winner Julien Alfred in the 100m female competition at Paris 2024 Olympic games.(© Getty Images)}
    \label{fig:input-img}
\end{wrapfigure}

We adopt a comprehensive three-tiered evaluation framework designed to rigorously assess our model's performance, and validate the integrity of the XKGs. The evaluation process encompasses: (i) logical validation of the triples, encompassing both syntactic correctness and proper anchoring to pre-existing nodes in the Base Graph; (ii) evaluation of foundational ontology alignment adequacy, and (iii) human assessment of the plausibility and adequacy of assertions within the triples generated for each heuristic.
%We provide a detailed examination of these three aspects in the subsequent sections.

To further ensure the robustness of our evaluation, we perform graph extension using an image captured after May 2024, shown in Figure \ref{fig:input-img}. This temporal selection is significant as it postdates the release of GPT-4o, thereby guaranteeing that the image is not part of the model's training dataset.

Due to space constraints, we present a detailed analysis and evaluation of a single knowledge extension instance, utilizing the image shown in Figure \ref{fig:input-img}, from the ``sport'' domain. Additional examples of knowledge extension, particularly focusing on ``politics'' and ``everyday life'' topics, are available in our dedicated GitHub repository\footnote{\url{https://github.com/StenDoipanni/XKG/tree/main/additional_use_cases}}.
We chose this image since it captures a moment of high emotional intensity. The athlete's expression clearly conveys what happened moments before and allows for informed speculation about subsequent events based on commonsense knowledge.
%We favored this dynamic scene over sports like e.g. swimming, where winning moments might be less visually expressive due to water coverage and equipment.

%\todo{Add information about the original graph number of triples, and the number of triples added per each heuristic}

\paragraph{\textbf{Base Graph}}
Following the pipeline shown in Figure~\ref{fig:pipeline},
we give as input the image shown in Figure \ref{fig:input-img} to GPT-4o, and we get the textual description shown in Box 1. To provide an example, highlighted in red is the anchoring text for Factual Impact knowledge, while in blue is the anchoring text for Moral Value-driven coercions.

\begin{tcolorbox}[basestyle, title=Box 1 - GPT Textual Description]
\footnotesize
\basetext{Jubilant female athlete celebrating victory on track, wearing Saint Lucia uniform with ``ALFRED'' on jersey, arms outstretched in triumphant pose, beaming with joy and exhilaration, fellow competitors visible behind her still finishing race, \factualtext{crowded stadium with cheering spectators in background}, American flag visible, symbolic moment of personal and national pride, \moraltext{representation of hard work and dedication paying off}, embodiment of Olympic spirit and international competition, capturing raw emotion and thrill of athletic achievement at highest level, inspiring scene of human potential realized.}
\label{box:gpt-text}
\end{tcolorbox}

The textual description is then passed to Text2AMR2FRED to produce the RDF graph, available as additional material on GitHub\footnote{Download and open in the browser this file: \url{https://github.com/StenDoipanni/XKG/blob/main/resources/graphs/base-graph.html}}.

\begin{figure}
    \centering
    \includegraphics[width=\linewidth]{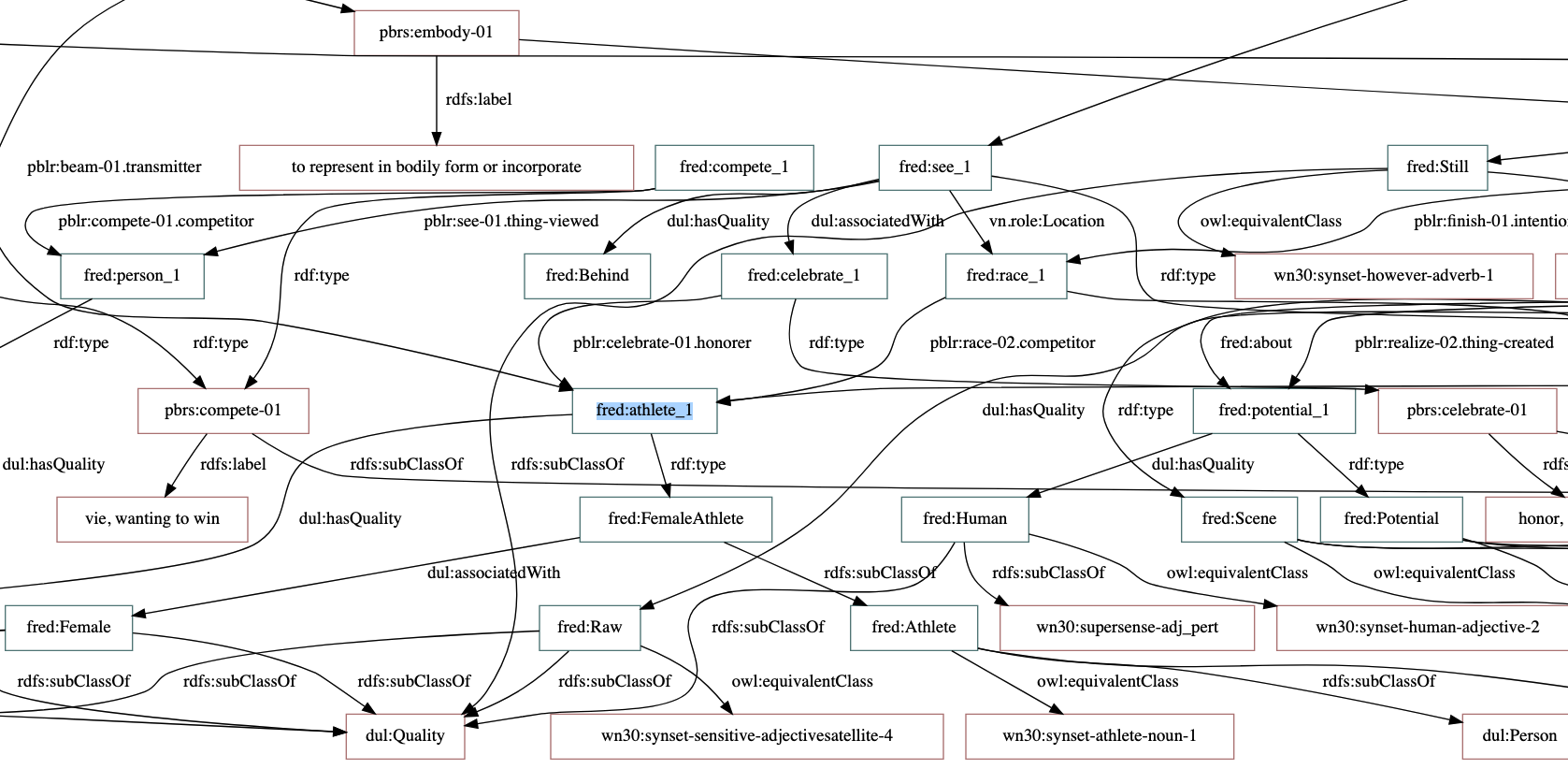}
    \caption{An excerpt of the AMR2FRED RDF graph.}
    \label{fig:amr2fred-excerpt}
\end{figure}

Figure \ref{fig:amr2fred-excerpt} shows an excerpt of the RDF graph obtained, in which it is possible to see the alignments to WordNet and PropBank, and to the DOLCE foundational ontology.
Box 2 shows an excerpt of the Base Graph, in which it is possible to see how e.g. the individual node \texttt{fred:athlete\_1} is aligned to DOLCE \texttt{dul:Person} and WordNet \texttt{wn:synset-athlete-noun-1}.
Furthermore, at the bottom of Box 2 there is a natural language transposition of the triples above. In red the anchoring points for Factual Impact knowledge (among others), and in blue the anchor for knowledge related to Moral Value-driven coercion.

The Base Graph contains $293$ OWL axioms (non-structural statements) as shown in Table \ref{tab:bg-stats}, which correspond to $1436$ RDF triples when serialized (including type declaration, label annotations, etc.). Out of these axioms, $21$ are equivalence axioms, expressed via the \texttt{owl:equivalentClass} property.

\begin{tcolorbox}[basestyle, title=Box 2 - Base graph]
\footnotesize
\basetext{fred:Athlete rdfs:subClassOf dul:Person,}\\
\indent \hspace{5mm}\basetext{wn30:supersense-noun\_person ;}\\
\indent \hspace{5mm}\basetext{owl:equivalentClass wn30:synset-athlete-noun-1 .}\\
\\
\basetext{fred:celebrate\_1 a pbrs:celebrate-01 ;}\\
\indent \hspace{5mm}\factualtext{vn.role:Location fred:track\_1 ;}\\
\indent \hspace{5mm}\moraltext{pblr:celebrate-01.honored fred:win\_1 ;}\\
\indent \hspace{5mm}\basetext{pblr:celebrate-01.honorer fred:athlete\_1 .}\\
\\
\basetext{The athlete is a Person. \factualtext{The gesture of celebration takes place on the track.} \moraltext{The victory is what is celebrated}, the athlete is the one celebrating.}
\end{tcolorbox}

\noindent Considering semantic web resources, the most retrieved are WordNet, PropBank and DOLCE: $33$ entities from WordNet are retrieved, regarding both physical entities such as \texttt{wn:synset-flag-noun-1}, \\ \texttt{wn:synset-uniform-noun-1} and \texttt{wn:synset-athlete-noun-1}, and more abstract ones such as \texttt{wn:synset-joy-noun-1}, and \\\texttt{wn:synset-international-adjective-1}.

\begin{table}[ht]
    \centering
    \small
    \setlength{\tabcolsep}{0.3em}
    \begin{tabular}{@{}lccccccl@{}} % Use @{} to remove left/right padding
        \toprule
        Graph & Axioms & WordNet & PB Roles & PB Frames & VN Roles & D0 & DUL \\ 
        \midrule
        Base Graph & 293 & 33 & 23 & 20 & 1 & 4 & 10 \\
        \bottomrule
    \end{tabular}
    \caption{Triples generated for the Base Graph detailed with origin of the arches and nodes.}
    \label{tab:bg-stats}
\end{table}

From PropBank we have $23$ local roles, namely instantiations of specific roles of a PropBank frame, including triples like:

\begin{quote}
    \texttt{fred:wear\_1} \texttt{pblr:wear-01.clothing} \texttt{fred:uniform\_1;} \\\texttt{pblr:wear-01.wearer} \texttt{fred:athlete\_1 .} 
\end{quote}

which means that the node ``wear\_1'' (namely an occurrence of wearing retrieved by FRED) takes as ``clothing'' role the node ``uniform\_1'' and as ``wearer'' role, the node ``athlete\_1''.
As for PropBank frames, we have $20$ distinct entities, among others: \texttt{pbrs:win-01}, \texttt{pbrs:achieve-01}, and \texttt{pbrs:celebrate-01}.
Finally, we retrieve $4$ entities from DOLCE 0:\texttt{d0:CognitiveEntity}, \texttt{d0:Event}, \texttt{d0:Characteristic}, and \\\texttt{d0:Activity}; and $10$ resources from DOLCE Ultralite, of which three are properties: \texttt{dul:associatedWith}, \texttt{dul:hasMember}, and \\\texttt{dul:hasQuality}, and $7$ are entities, such as \texttt{dul:Person}, \\\texttt{dul:Situation}, and \texttt{dul:InformationEntity}.
All the SPARQL queries to investigate the graphs are available as additional materials on the GitHub\footnote{\url{https://github.com/StenDoipanni/XKG/tree/main/resources/sparql-queries}}.

\paragraph{\textbf{Extended Knowledge Graphs}}
We summarize here some numbers about the 11 XKGs generated with the heuristics; all the graphs are available on GitHub\footnote{\url{https://github.com/StenDoipanni/XKG/tree/main/resources/graphs}}.

\begin{table}[ht]
    \centering
    \small
    \setlength{\tabcolsep}{0.3em}
    \begin{tabular}{@{}lcccccl@{}} % Use @{} to remove left/right padding
        \toprule
        Graph & Axioms & WordNet & PB Roles & PB Entities & OP & DP \\ 
        \midrule
        Presuppositions & 32 & - & - & 3 & - & 11 \\
        Conversational Implicatures & 36 & 6 & - & - & 14 & - \\
        Factual Impact & 13 & 5 & - & - & 3 & - \\
        Image Schemas & 63 & 11 & - & - & 1 & - \\
        Metonymic Coercion & 26 & - & 5 & 5 & 7 & - \\
        Moral Value-driven Coercion & 12 & - & 2 & 1 & 3 & - \\
        Symbolic Coercion & 15 & 7 & - & - & 1 & - \\
        Event Sequences & 15 & - & - & - & 1 & - \\
        Causal Relations & 16 & 1 & - & - & 1 & - \\
        Implied Future Events & 14 & - & 1 & 1 & 2 & - \\
        Potential Non-events & 23 & - & 5 & 4 & 5 & - \\
        \bottomrule
    \end{tabular}
    \caption{Triples generated for the Base Graph and the 11 heuristics graphs, including axioms count, presence of WordNet entities, PropBank Roles, PropBank frames, and newly introduced Object Properties or Data Properties.}
    \label{tab:xkg-stats}
\end{table}

%\todo{Add small Box viz of some of the XKGs}

Table \ref{tab:xkg-stats} omits the columns for D0 and DUL, since most of XKGs do not declare alignments to DOLCE classes. Two notable exceptions are: the \texttt{dul:precedes} property, used to state sequential order of entities, which is extensively present in Presuppositions as well as Event Sequences, and Implied Future Events XKGs; and the Image Schema XKG, which presents $14$ alignments of image schemas to DUL classes; among them: ``Balance'' subClass of \texttt{dul:Quality}, ``Collection'' subClass of \texttt{dul:Collection}, ``Cicle'' subClass of \texttt{dul:Process}, ``Path'' subClass of \texttt{dul:SpaceRegion}, and ``Container'' as subClass of \texttt{dul:PhysicalObject}.

Box 3 shows and exceprt of triples generated in the Factual Impact XKG.
We discuss the plausibility and correctness of alignments and triples in the next sections, via a three-tier evaluation.

\begin{tcolorbox}[basestyle, title= Box 3 - Factual Impact, coltitle=factualcolor]
\footnotesize
\basetext{fred:athlete\_1} \factualtext{impact:hasExpectedEmotion impact:Joy,} \\
\indent \hspace{5mm}\factualtext{impact:Pride ;}\\
\indent \hspace{5mm}\factualtext{impact:hasExpectedPhysicalState impact:Exhilaration ;}\\
\indent \factualtext{ impact:hasExpectedSocialImpact impact:NationalRecognition.}\\
\\
\basetext{The athlete} \factualtext{is expected to feel joy and pride, it is expected to be ecstatic, and to receive some form of national recognition} \basetext{for winning.}\\
\end{tcolorbox}

A particularly salient aspect of XKGs is the introduction of novel properties, as delineated in the final two columns of Table \ref{tab:xkg-stats}. These columns, labeled OP and DP, represent newly introduced ``Object Properties'' and ``Datatype Properties'' respectively, offering insight into the ontological expansion and semantic extension of each knowledge graph.

Table \ref{tab:xkg-stats} presents an overview of the composition of XKGs. The graphs exhibit significant variation in their structural complexity and semantic richness, reflecting the diverse nature of the conceptual domains being modeled. The ``Axioms'' column reveals considerable differences in the logical foundations of each graph, ranging from $12$ axioms in the Moral Value-driven Coercion graph to a notable $63$ in the Image Schemas graph. This substantial difference can be traced back to the nature of the reference image being a sport scene, which likely fosters a richer generation of triples, and representation of knowledge related to spatial relations, bodily movements, and physical interactions—concepts central to Image Schemas. The integration of external lexical-semantic resources is evident, with WordNet entities being prominently featured in several graphs, particularly in Conversational Implicatures ($6$) and Image Schemas ($11$). PropBank roles and entities are less frequently incorporated, with the Event Sequences graph showing the highest utilization ($5$ PB Entities), possibly indicating a focus on action-oriented semantics in depicting the sequential nature of plausible events. The graphs also exhibit varying degrees of expressiveness, as indicated by the introduction of new Object Properties (OP) and Datatype Properties (DP). The Conversational Implicatures graph stands out in this regard, introducing $14$ new Object Properties, among which we see \texttt{:hasVictory}, \texttt{:hasEmotion}, \texttt{:hasNationality}, and \texttt{:hasSignificance}, showing several semantic layers collapsed in the same XKG, including a huge amount of implicit knowledge that could be further unpacked. Similarly, the Presuppositions graph adds $11$ new Data Properties. It is noteworthy the fact that all of them takes as object a boolean `True' vs `False', and all of them takes as value `True', pointing in the direction that, in lack of a certain information, e.g. the exact date in which the stadium was built, it is still possible to represent on the graph this information introducing an axiom like: 

\begin{quote}
    \texttt{fred:stadium\_1} \texttt{fred:wasBuiltBefore} \texttt{true}.
\end{quote}

This heterogeneity in graph composition not only reflects the diverse requirements for semantic expressivity across different areas of knowledge representation but also underscores how different domain-related images e.g. sports-centric vs political, could influence the depth and breadth of implicit knowledge extraction.

\paragraph{\textbf{Logical Integrity and Foundational Ontologies Compliance}}
To ensure the structural integrity and logical consistency of all XKGs, we use the Hermit $1.4.3.456$ reasoner \cite{glimm2014hermit} as part of our evaluation framework, as well as OOPS! - OntOlogy Pitfall Scanner \cite{poveda2014oops}.

\begin{wrapfigure}{l}{0.4\linewidth}
    \centering
    \includegraphics[width=\linewidth]{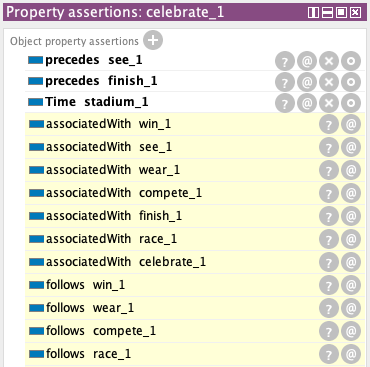}
    \caption{Protégé vizualisation of inferences obtained running Hermit $1.2.3.456$ on the Event Sequences extended graph.}
    \label{fig:time-inference}
\end{wrapfigure}

The reasoner's check is meant to verify soundness and consistency of the newly introduced LLM-generated triples, thereby ensuring that the extended knowledge remains sound and usable for downstream applications and inference tasks. In particular, we use Hermit via Protégé $5.5.0$ interface and OOPS! web interface\footnote{\url{https://oeg.fi.upm.es/index.php/en/technologies/292-oops/index.html}}.
Passing each and every XKG graph to OOPS!, the pitfall scanner yields minor issues for all of them, mainly related to the absence of metadata such as \texttt{rdfs:comment} describing entities.

Occasional instances of hallucination can been observed, particularly in the application of prefixes. An example occurs in the Metonymic coercion file, where the prefix ``pbrs'' (denoting PropBank Role Set, which PropBank utilized for frame-like structures) is erroneously employed instead of the correct prefix ``pblr'' (PropBank Local Role), which is the appropriate designation for occurrences of PropBank roles. Such inconsistencies, while minor, underscore the importance of rigorous post-processing and validation in ensuring the accuracy and reliability of the model's semantic annotations, as described in Section \ref{sec:ongoing}.

For soundness validation, we checked each XKG importing both the Base Graph and DOLCE Zero, in order to ensure complete coherence with the AMR2FRED original graph, as well as the DOLCE foundational ontology.

An isolated instance of inconsistency is identified in the Metonymic Coercions heuristic graph, suggesting the fact this XKG could present problems on several sides. This anomaly manifested as a conflicting chain of subsumptions ($athlete\_1 \rightarrow Athlete \rightarrow Person \rightarrow Agent \rightarrow Object$) wherein a single entity is erroneously classified as both an Object and an Event. Specifically, `athlete\_1' is incorrectly categorized as an instance of `wn:cheer-01', rather than being appropriately designated as an agent of the cheering action. It is noteworthy that while the \texttt{d0:Event} class is deprecated, we have opted to retain it in our consistency verification process due to its relevance in identifying misalignments.

Other than checking for inconsistencies, found only in the Metonymic Coercion XKG, as mentioned above, reasoning over XKGs with the import of Base Graph and DOLCE 0, yields relevant inferences, as shown in Figure \ref{fig:time-inference}. The Event Sequences XKG, in fact, making use of the \texttt{d0:precedes} property, allows to infer the order of events thanks to the transitivity of the property, positioning them in sequential order not explicitly stated. In this case, as shown in Figure \ref{fig:time-inference}, the occurrence of celebration is inferred as sequentially posterior to the racing action, the competition event, the wearing action, and the winning event.

\paragraph{\textbf{XKGs Human Evaluation}}
To ensure the quality and reliability of the generated triples, we conduct a comprehensive human evaluation process. The validation was performed by 5 annotators, all of whom possess proficiency in Resource Description Framework (RDF) and Turtle syntax, but are not domain experts across all 11 heuristics considered. Each triple produced in our extension undergoes scrutiny and is labeled using a 5-point Likert scale, where 1 represents ``Not at all plausible/adequate'' and 5 indicates ``Completely plausible/adequate''. This approach allows for a nuanced assessment of the triples' validity and relevance within their respective domains. By employing annotators with RDF expertise but varying levels of domain-specific knowledge, we aims to balance technical accuracy with a generalist perspective, mirroring real-world scenarios where RDF data may be consumed by users with diverse backgrounds.

%\subsection{XKGs Validation}
%For the sports competition image, we selected a photograph from the Paris 2024 Olympic Games. This choice serves multiple purposes: it's outside the model's training set, and 

%\paragraph{\textbf{Heuristics Enrichment Triples}}

%Here you see some triples from the ``Base Graph'' namely the graph that is given as output from original image description only.

%Factual Impact enrichment provided triples about ...

%And then moral values enrichment introduced ...

%\begin{tcolorbox}[basestyle, title=MORAL VALUES, coltitle=moralcolor]
%\footnotesize
%\moraltext{fred:assert\_1 a pbrs:assert-02 ;}\\
%\hspace{5mm}\moraltext{mor:evokes mor:PowerDemonstration ;}\\
%\hspace{5mm}\moraltext{pblr:assert-02.agent fred:figure\_2 ;}\\
%\hspace{5mm}\moraltext{pblr:assert-02.topic fred:power\_1 .}\\
%\\
%\basetext{The figure assertion is a \moraltext{demonstration of power}.}
%\end{tcolorbox}

%\subsection{XKG Logical Validation}
%\label{sec:xkg-oops}

%As shown in the mTAF Web UI 

%\subsection{XKG Foundational Ontology Alignment Validation}
%\label{sec:xkg-dolce}

%\subsection{XKG Enrichment Human Validation}
%\label{sec:xkg-human}

%Each triple is annotated by 5 raters on a 5 points Likert scale where 1 = ``Not at all plausible/adequate'' and 5 = ``Totally plausible/adequate''.
Furthermore, to ensure the reliability and consistency of our ratings, we employs multiple statistical measures. Inter-rater agreement is calculated to assess the overall concordance among raters. Krippendorff's alpha is chosen for its ability to handle ordinal data and accommodate multiple raters, providing a robust measure of reliability. Cohen's Kappa, while typically used for binary ratings, is adapted to evaluate pairwise agreement between raters. Mean ratings are computed to provide a central tendency measure of the perceived quality of the generated triples. Finally, we calculate the standard deviation to quantify the dispersion of ratings, to get further insight into the consistency or variability of assessments across raters.

\paragraph{\textbf{Mean Ratings and Standard Deviation}}

\begin{figure}
    \centering
    \includegraphics[width=\linewidth]{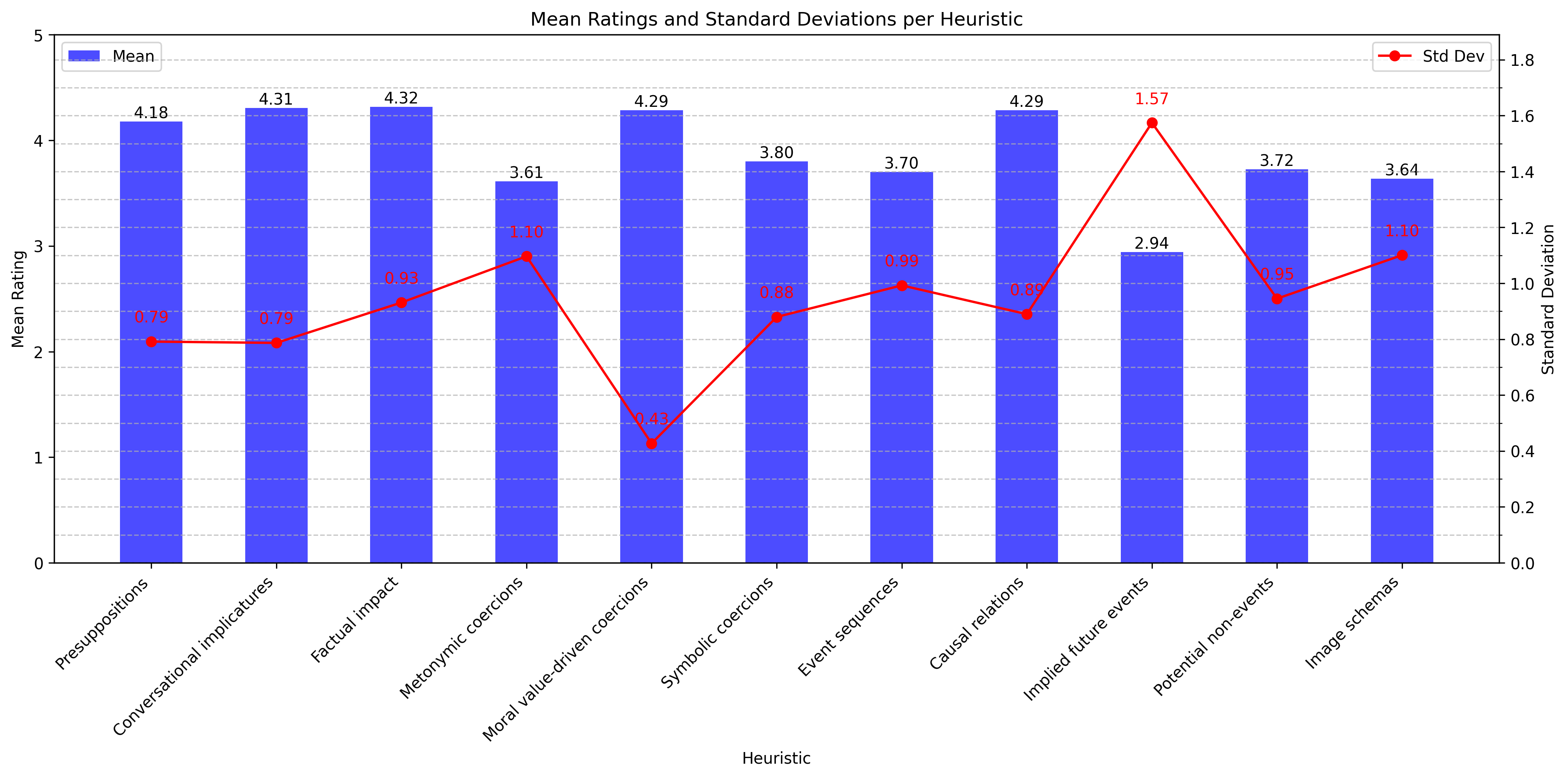}
    \caption{Mean Ratings and Standard Deviation per Heuristics.}
    \label{fig:mean-ratings}
\end{figure}

It is important to highlight that, given the 5 point Likert scale, almost all the heuristics overall passed the threshold of 3, with the exception of Implied Future Events, which still shows a rating of $2.94$. This means that, overall, the XKGs present at least ``fairly plausible'' knowledge extension for all the domains.
This is \textit{per se} a remarkable achievement, given the disparity among the domains, which coupled with symbolic reasoning inference regarding e.g. sequences of events, mentioned above, opens to very promising further neurosymbolic methodology exploration.

Furthermore, the analysis of mean ratings and standard deviations across various heuristics shown in Figure \ref{fig:mean-ratings} reveals interesting patterns in the performance and consistency of different evaluation criteria. Factual Impact, Conversational Implicatures, and Moral Value-driven Coercions emerge as the top-performing heuristics, with mean ratings exceeding $4.29$ on a 5-point scale. This suggests a high degree of plausibility or adequacy in these areas. Conversely, Potential Non-Events and Image Schemas received the lowest mean ratings ($2.94$ and $3.64$, respectively), indicating potential areas for improvement in the language model's output. Notably, the standard deviations exhibit considerable variation, with Implied Future Events showing the highest variability (SD = $1.57$) and Moral Value-driven Coercions demonstrating the most consistency (SD = $0.47$). This disparity in standard deviations highlights the varying levels of agreement among raters across different heuristics. The data underscores the need for targeted refinements in certain aspects of the language model's performance, particularly in areas with lower mean ratings or higher standard deviations, to enhance the overall quality and consistency of generated content, which is discussed in Section \ref{sec:ongoing}.
There might be a possible ``remoteness effect'' when users evaluate uncertain (as with future events) or very abstract (as with image schemas) implicit knowledge, which should be further investigated by involving field experts.

\paragraph{\textbf{Mean Scores per Annotator}}

\begin{figure}
    \centering
    \includegraphics[width=\linewidth]{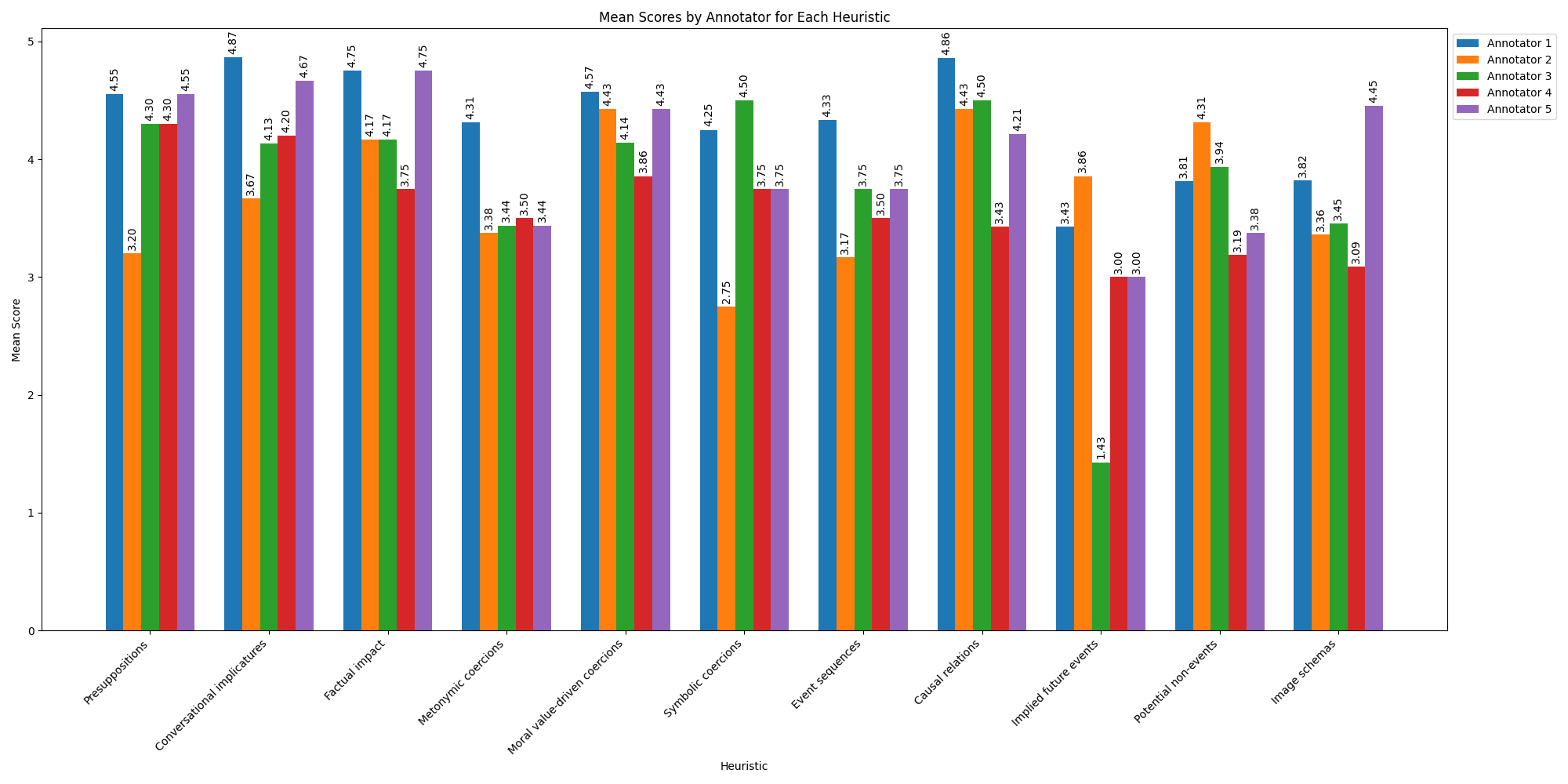}
    \caption{Mean Scores by Annotator for Each Heuristic.}
    \label{fig:annotator-scores}
\end{figure}

The analysis of mean scores by annotator for each heuristic, shown in Figure \ref{fig:annotator-scores}, reveals significant insights into the evaluation process and possibly the nature of the heuristics themselves. Notably, there is considerable variation in scoring patterns across annotators, suggesting potential differences in interpretation or application of the evaluation criteria. Annotator 1 consistently provided higher scores across most heuristics, particularly for Conversational Implicatures ($4.87$) and Factual Impact ($4.75$), while Annotator 3 tended to score more conservatively, especially for Implied Future Events ($1.43$). The heuristics of Factual Impact and Causal Relations demonstrated relatively high consensus among annotators, with most scores clustering above $4.0$, indicating their robustness and clarity. Conversely, Implied Future Events and Potential Non-events exhibited the widest disparity in scores, ranging from $1.43$ to $3.86$ and $3.19$ to $4.31$ respectively, highlighting areas where the evaluation criteria may benefit from refinement. This variability underscores the subjective nature of certain heuristics and emphasizes the importance of clear guidelines and calibration sessions in future annotation tasks to enhance inter-rater reliability and the overall validity of the evaluation process.

Furthermore, the scoring patterns reveal distinct tendencies among the five annotators, showcasing varying levels of ``generosity'' in their evaluations:
Annotator 1 emerges as the most generous evaluator, consistently providing the highest scores across most heuristics. This is particularly evident in Conversational Implicatures ($4.87$), Factual Impact ($4.75$), and Moral Value-driven Coercions ($4.57$). Their scores are frequently at least $0.5$ points higher than the next highest annotator, suggesting a more lenient interpretation of the evaluation criteria.
Annotator 5 appears to be the second most generous, often providing scores close to, but slightly below, Annotator 1. They show particular generosity in Presuppositions ($4.55$) and Factual Impact ($4.75$), matching Annotator 1 in the latter.
Annotator 2 demonstrates a more moderate approach, typically scoring in the middle range compared to other annotators. However, they show higher scores for Causal Relations ($4.43$) and Potential Non-Events ($4.31$), indicating possible areas of expertise or confidence.
Annotator 4 tends to be more conservative in their scoring, often providing lower scores than Annotators 1, 2, and 5. This is particularly noticeable in heuristics like Metonymic Coercions ($3.44$) and Implied Future Events ($3.00$). However, they align more closely with others on some heuristics like Presuppositions ($4.30$).
Annotator 3 emerges as the most stringent evaluator overall. They consistently provide the lowest scores across multiple heuristics, most notably in Implied Future Events ($1.43$) and Image Schemas ($3.09$). This suggests a more critical approach to the evaluation process or possibly a stricter interpretation of the scoring criteria.

\paragraph{\textbf{Agreement Measures}}

\begin{figure}
    \centering
    \includegraphics[width=\linewidth]{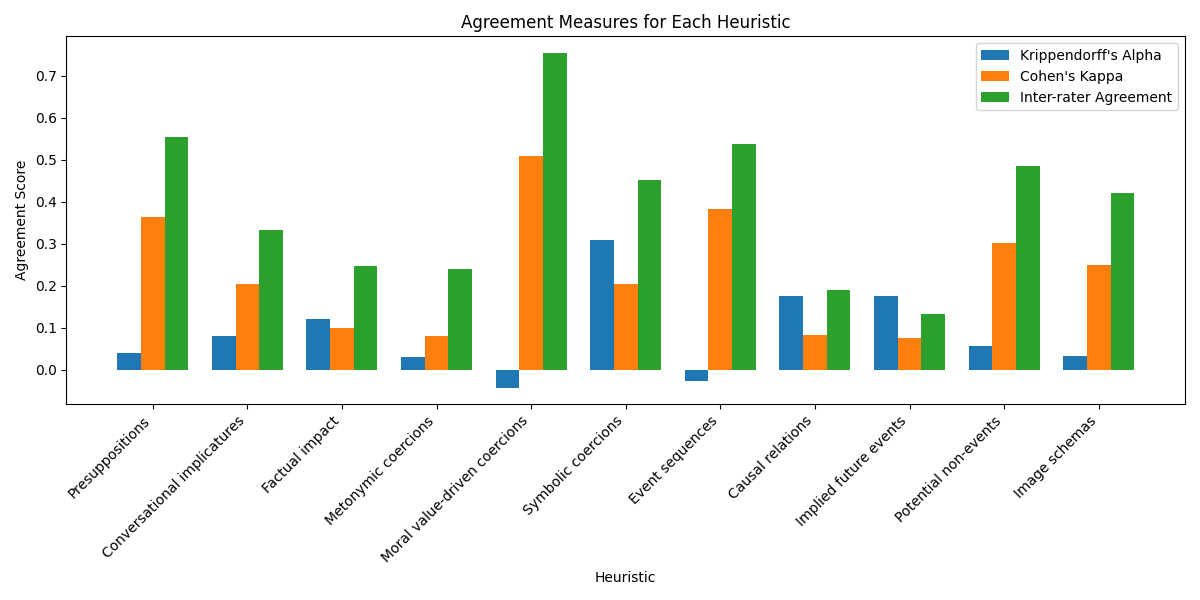}
    \caption{Agrement measures}
    \label{fig:agreement-measures}
\end{figure}

%The analysis of agreement measures across heuristics reveals significant variations in inter-rater reliability, providing crucial insights into the robustness of each evaluation criterion. 
Figure \ref{fig:agreement-measures} shows how Moral Value-driven Coercions demonstrate the highest overall agreement, with an inter-rater agreement of $0.75$ and a Cohen's Kappa of $0.51$, suggesting strong consistency among raters for this heuristic. Conversely, Implied Future Events show the lowest agreement across all measures, indicating a potential need for refinement in its definition or evaluation criteria. Notably, Krippendorff's Alpha consistently yields lower values compared to other measures, with several heuristics showing negative values, particularly for Metonymic Coercions and Moral Value-driven Coercions. This discrepancy between Krippendorff's Alpha and other measures warrants further investigation, as it may indicate sensitivity to specific patterns in the data or potential limitations in applying this metric to the current evaluation framework. The generally moderate to low agreement scores across most heuristics, especially for measures like Factual Impact and Conversational Implicatures, underscore the challenges in achieving consistent evaluations for potentially conceptually complex phenomena. 

Following AAAI-20 talk by Henry Kautz\footnote{\url{https://www.youtube.com/watch?v=_cQITY0SPiw}}, these results confirm that the usage of LLMs in our hybrid pipeline moves significant steps not towards expert reasoning, but mostly towards commonsense reasoning.

%These findings suggest possibly the need for more rigorous rater training, clearer evaluation guidelines, and possibly the refinement of certain heuristic definitions to enhance the reliability and validity of the assessment process in future studies.

%% file: 5_ongoing.tex
In our ongoing efforts to refine and enhance the generation of XKGs, several specific improvements are being explored. Prompt refinement for in-context learning has emerged as a crucial area for development, particularly in light of insights gained from image schemas. Our current approach utilizes a standard template, but evidence suggests that more adapted heuristic-specific prompts could yield superior results. 
%This aligns with findings from other studies demonstrating improved model performance when employing more domain-expert tailored prompts.

Another area of focus is the refinement of property assignments. Currently, many triples are generated using the broad \texttt{dul:associatedWith} top property. Efforts are underway to specialize this property further, aligning newly introduced properties with DOLCE Zero. While this integration may increase the risk of ontological inconsistencies, it also promises enhanced inferential capabilities.
%exemplifying the complex trade-offs inherent in ontology alignment and knowledge base enrichment.

We are also reevaluating our validation methodology. The current system\footnote{Available here:\url{https://github.com/StenDoipanni/XKG/blob/main/resources/XKG-human-validation-form.pdf}}, which informs annotators of the heuristic definition they are validating, may be influencing results. An alternative approach, such as presenting uncontextualized triples for evaluation, could potentially yield different outcomes in human assessment.

Lastly, we are exploring alignment with domain-specific ontologies, particularly those focused on Image Schemas \cite{de2022imageschemanet}, moral and cultural values \cite{de2022basic}, and other cognitive entities such as emotions. This could involve refining prompts to incorporate either complete ontology schemas (when feasible) or at least excerpts of top taxonomy classes, potentially leading to more nuanced and domain-specific knowledge.

Future work will focus on several key areas to enhance the capabilities and applicability of our graph enrichment process. A primary objective is the topicalization of enrichment when starting from images, which involves localizing triples related to specific heuristics to particular portions of an image. This can be achieved through the utilization of existing labeled repositories such as Visual Genome or ImageNet, or by implementing object-scene recognition algorithms that disambiguate to WordNet synsets before proceeding with triple generation.

Addressing the challenges of Wikidata entry alignment is another crucial area of development. While AMR2FRED proves reliable for the Base Graph, LLM-based alignment has shown significant inconsistencies, highlighting issues with precise information retrieval and hallucinations. A proposed solution involves using the ``wd:'' prefix with entity labels, followed by script-based verification using SPARQL engines like Qlever \cite{bast2017qlever}. This method, incorporating CamelCase string parsing, shows promise for specific named entities but may require further refinement for general conceptual entities.

Although our method is model-agnostic, practical results vary across different LLMs. Current implementation relies on state-of-the-art models requiring proprietary APIs. Future efforts will focus on refining prompts and segmenting processes to enable the use of smaller yet efficient models like Phi 3.5, potentially broadening accessibility.
%and reducing dependency on costly resources.

Lastly, we envision expanding our service to accommodate user-defined prompts, opening up a vast array of possibilities for knowledge graph extension. This could include specialized applications such as e.g., color coercions for environmental elements (we assume that if there is a portion of image which represents the sea, or a forest, or snow, etc. there could be a specific color palette) or e.g. identifying potential safety hazards in images (this could have relevant impact in social robotics). The potential applications are diverse and hold significant relevance across various domains, underscoring the expansive capabilities of our approach in knowledge representation and reasoning.

%% file: 6_conclusions.tex
Our research presents a significant step forward in the development of a hybrid neurosymbolic method for grounded world models that can effectively integrate multiple layers of knowledge, bridging the gap between the pattern-matching reactive capabilities of LLMs, and the structured reasoning applied on Knowledge Bases and traditional symbolic reasoning. By leveraging LLMs as repositories of implicit commonsense knowledge rather than expert systems, we have demonstrated a novel approach to knowledge base extension that is both agile and comprehensive.

These results also add to a growing corpus of results about the hidden correspondence of massive linguistic knowledge to the structure of human environments, be them physical, social, cognitive or purely abstract.

The multi-tiered evaluation framework we employed, encompassing logical validation, foundational ontology alignment, and human assessment, provides strong evidence for the efficacy of our approach and future promising refinements. The high plausibility ratings across most heuristics, particularly in areas such as Factual Impact, Conversational Implicatures, and Moral Value-driven Coercions, underscore the potential of our method to capture nuanced semantic information that goes beyond simple fact retrieval.

However, the challenges revealed in our evaluation, particularly in areas like Implied Future Events and Image Schemas, point to important directions for future work. The variability in inter-rater agreement across different heuristics highlights the complexity of evaluating implicit knowledge and the need for continued refinement of our methods and evaluation criteria, even if uncertainty (for future events) and abstractness (for image schemas) of implicit knowledge are probably good reasons for lower agreement.

The neuro-symbolic nature of our approach, combining the strengths of neural networks and symbolic reasoning, opens up new possibilities for systems that can flexibly adapt to novel contexts while maintaining the ability to perform structured inference. This hybrid architecture addresses longstanding challenges in AI, such as the Frame problem, by providing a mechanism for dynamically integrating diverse knowledge sources and reasoning patterns.

Looking ahead, our work lays the foundation for several promising research directions. The potential for topicalization of enrichment on images, improved alignment with domain-specific ontologies, and the exploration of user-defined prompts for specialized applications all represent exciting avenues for extending the capabilities of our system.

In conclusion, our research contributes to the broader goal of developing systems capable of human-like reasoning and contextual understanding relevant in several areas ranging from natural language processing and computer vision to complex problem-solving and decision-making in real-world scenarios.

%% file: 0_main.bbl
@inproceedings{meyer2023llm,
  title={Llm-assisted knowledge graph engineering: Experiments with chatgpt},
  author={Meyer, Lars-Peter and Stadler, Claus and Frey, Johannes and Radtke, Norman and Junghanns, Kurt and Meissner, Roy and Dziwis, Gordian and Bulert, Kirill and Martin, Michael},
  booktitle={Working conference on Artificial Intelligence Development for a Resilient and Sustainable Tomorrow},
  pages={103--115},
  year={2023},
  organization={Springer Fachmedien Wiesbaden Wiesbaden}
}

@article{gangemi2016kbs,
  author       = {Aldo Gangemi and
                  Diego Reforgiato Recupero and
                  Misael Mongiov{\`{\i}} and
                  Andrea Giovanni Nuzzolese and
                  Valentina Presutti},
  title        = {Identifying motifs for evaluating open knowledge extraction on the
                  Web},
  journal      = {Knowl. Based Syst.},
  volume       = {108},
  pages        = {33--41},
  year         = {2016},
  url          = {https://doi.org/10.1016/j.knosys.2016.05.023},
  doi          = {10.1016/J.KNOSYS.2016.05.023},
  timestamp    = {Fri, 27 Mar 2020 08:38:08 +0100},
  biburl       = {https://dblp.org/rec/journals/kbs/GangemiRMNP16.bib},
  bibsource    = {dblp computer science bibliography, https://dblp.org}
}

@article{zhu2024llms,
  title={Llms for knowledge graph construction and reasoning: Recent capabilities and future opportunities},
  author={Zhu, Yuqi and Wang, Xiaohan and Chen, Jing and Qiao, Shuofei and Ou, Yixin and Yao, Yunzhi and Deng, Shumin and Chen, Huajun and Zhang, Ningyu},
  journal={World Wide Web},
  volume={27},
  number={5},
  pages={58},
  year={2024},
  publisher={Springer}
}

@inproceedings{sen2023knowledge,
  title={Knowledge graph-augmented language models for complex question answering},
  author={Sen, Priyanka and Mavadia, Sandeep and Saffari, Amir},
  booktitle={Proceedings of the 1st Workshop on Natural Language Reasoning and Structured Explanations (NLRSE)},
  pages={1--8},
  year={2023}
}

@article{edge2024local,
  title={From local to global: A graph rag approach to query-focused summarization},
  author={Edge, Darren and Trinh, Ha and Cheng, Newman and Bradley, Joshua and Chao, Alex and Mody, Apurva and Truitt, Steven and Larson, Jonathan},
  journal={arXiv preprint arXiv:2404.16130},
  year={2024}
}

@article{yao2019kg,
  title={KG-BERT: BERT for knowledge graph completion},
  author={Yao, Liang and Mao, Chengsheng and Luo, Yuan},
  journal={arXiv preprint arXiv:1909.03193},
  year={2019}
}

@article{lu2020utilizing,
  title={Utilizing textual information in knowledge graph embedding: A survey of methods and applications},
  author={Lu, Fengyuan and Cong, Peijin and Huang, Xinli},
  journal={IEEE Access},
  volume={8},
  pages={92072--92088},
  year={2020},
  publisher={IEEE}
}

@inproceedings{tiwari2022preface,
  title={Preface for the International Workshop on Knowledge Graph Generation from Text},
  author={Tiwari, Sanju and Mihindukulasooriya, Nandana and Osborne, Francesco and Kontokostas, Dimitris and D’Souza, Jennifer and Kejriwal, Mayank and others},
  booktitle={CEUR WORKSHOP PROCEEDINGS},
  volume={3184},
  year={2022},
  organization={CEUR}
}

@article{poveda2014oops,
  title={Oops!(ontology pitfall scanner!): An on-line tool for ontology evaluation},
  author={Poveda-Villal{\'o}n, Mar{\'\i}a and G{\'o}mez-P{\'e}rez, Asunci{\'o}n and Su{\'a}rez-Figueroa, Mari Carmen},
  journal={International Journal on Semantic Web and Information Systems (IJSWIS)},
  volume={10},
  number={2},
  pages={7--34},
  year={2014},
  publisher={IGI Global}
}

@inproceedings{fathallah2024neon,
  title={NeOn-GPT: A Large Language Model-Powered Pipeline for Ontology Learning},
  author={Fathallah, Nadeen and Das, Arunav and De Giorgis, Stefano and Poltronieri, Andrea and Haase, Peter and Kovriguina, Liubov},
  booktitle={The Extended Semantic Web Conference},
  year={2024}
}

@article{glimm2014hermit,
  title={HermiT: an OWL 2 reasoner},
  author={Glimm, Birte and Horrocks, Ian and Motik, Boris and Stoilos, Giorgos and Wang, Zhe},
  journal={Journal of automated reasoning},
  volume={53},
  pages={245--269},
  year={2014},
  publisher={Springer}
}

@inproceedings{paulheim2015serving,
  title={Serving DBpedia with DOLCE--more than just adding a cherry on top},
  author={Paulheim, Heiko and Gangemi, Aldo},
  booktitle={The Semantic Web-ISWC 2015: 14th International Semantic Web Conference, Bethlehem, PA, USA, October 11-15, 2015, Proceedings, Part I 14},
  pages={180--196},
  year={2015},
  organization={Springer}
}

@inproceedings{bevilacqua2021one,
  title={One SPRING to rule them both: Symmetric AMR semantic parsing and generation without a complex pipeline},
  author={Bevilacqua, Michele and Blloshmi, Rexhina and Navigli, Roberto},
  booktitle={Proceedings of the AAAI Conference on Artificial Intelligence},
  volume={35},
  number={14},
  pages={12564--12573},
  year={2021}
}

@article{wu2019scalable,
  title={Scalable zero-shot entity linking with dense entity retrieval},
  author={Wu, Ledell and Petroni, Fabio and Josifoski, Martin and Riedel, Sebastian and Zettlemoyer, Luke},
  journal={arXiv preprint arXiv:1911.03814},
  year={2019}
}

@inproceedings{meloni2017amr2fred,
  title={AMR2FRED, a tool for translating abstract meaning representation to motif-based linguistic knowledge graphs},
  author={Meloni, Antonello and Reforgiato Recupero, Diego and Gangemi, Aldo},
  booktitle={The Semantic Web: ESWC 2017 Satellite Events: ESWC 2017 Satellite Events, Portoro{\v{z}}, Slovenia, May 28--June 1, 2017, Revised Selected Papers 14},
  pages={43--47},
  year={2017},
  organization={Springer}
}

@inproceedings{gangemi2017amr2fred,
  title={AMR2FRED, A Tool for Translating Abstract Meaning Representation to Motif-Based Linguistic Knowledge Graph},
  author={Gangemi, Aldo and others},
  booktitle={Proceedings of the Extended Semantic Web Conference (ESWC2017)},
  pages={43--47},
  year={2017},
  organization={DEU}
}

@inproceedings{bevilacqua-navigli-2020-breaking,
    title = "Breaking Through the 80{\%} Glass Ceiling: {R}aising the State of the Art in Word Sense Disambiguation by Incorporating Knowledge Graph Information",
    author = "Bevilacqua, Michele  and Navigli, Roberto",
    booktitle = "Proceedings of the 58th Annual Meeting of the Association for Computational Linguistics",
    month = jul,
    year = "2020",
    address = "Online",
    publisher = "Association for Computational Linguistics",
    url = "https://www.aclweb.org/anthology/2020.acl-main.255",
    pages = "2854--2864"
}

@article{hamilton2022neuro,
  title={Is neuro-symbolic ai meeting its promises in natural language processing? a structured review},
  author={Hamilton, Kyle and Nayak, Aparna and Bo{\v{z}}i{\'c}, Bojan and Longo, Luca},
  journal={Semantic Web},
  number={Preprint},
  pages={1--42},
  year={2022},
  publisher={IOS Press}
}

@article{kautz2022third,
  title={The third ai summer: Aaai robert s. engelmore memorial lecture},
  author={Kautz, Henry},
  journal={Ai magazine},
  volume={43},
  number={1},
  pages={105--125},
  year={2022}
}

@inproceedings{gangemi2016framester,
  title={Framester: A wide coverage linguistic linked data hub},
  author={Gangemi, Aldo and Alam, Mehwish and Asprino, Luigi and Presutti, Valentina and Recupero, Diego Reforgiato},
  booktitle={Knowledge Engineering and Knowledge Management: 20th International Conference, EKAW 2016, Bologna, Italy, November 19-23, 2016, Proceedings 20},
  pages={239--254},
  year={2016},
  organization={Springer}
}

@inproceedings{belle2020symbolic,
  title={Symbolic logic meets machine learning: A brief survey in infinite domains},
  author={Belle, Vaishak},
  booktitle={International conference on scalable uncertainty management},
  pages={3--16},
  year={2020},
  organization={Springer}
}

@incollection{besold2021neural,
  title={Neural-symbolic learning and reasoning: A survey and interpretation 1},
  author={Besold, Tarek R and d’Avila Garcez, Artur and Bader, Sebastian and Bowman, Howard and Domingos, Pedro and Hitzler, Pascal and K{\"u}hnberger, Kai-Uwe and Lamb, Luis C and Lima, Priscila Machado Vieira and de Penning, Leo and others},
  booktitle={Neuro-Symbolic Artificial Intelligence: The State of the Art},
  pages={1--51},
  year={2021},
  publisher={IOS press}
}

@article{garcez2023neurosymbolic,
  title={Neurosymbolic AI: The 3 rd wave},
  author={Garcez, Artur d’Avila and Lamb, Luis C},
  journal={Artificial Intelligence Review},
  volume={56},
  number={11},
  pages={12387--12406},
  year={2023},
  publisher={Springer}
}

@article{lamb2020graph,
  title={Graph neural networks meet neural-symbolic computing: A survey and perspective},
  author={Lamb, Lu{\'\i}s C and Garcez, Artur and Gori, Marco and Prates, Marcelo and Avelar, Pedro and Vardi, Moshe},
  journal={arXiv preprint arXiv:2003.00330},
  year={2020}
}

@article{sarker2021neuro,
  title={Neuro-symbolic artificial intelligence},
  author={Sarker, Md Kamruzzaman and Zhou, Lu and Eberhart, Aaron and Hitzler, Pascal},
  journal={AI Communications},
  volume={34},
  number={3},
  pages={197--209},
  year={2021},
  publisher={IOS Press}
}

@article{yao2018learning,
  title={Learning to activate logic rules for textual reasoning},
  author={Yao, Yiqun and Xu, Jiaming and Shi, Jing and Xu, Bo},
  journal={Neural Networks},
  volume={106},
  pages={42--49},
  year={2018},
  publisher={Elsevier}
}

@article{yu2023survey,
  title={A survey on neural-symbolic learning systems},
  author={Yu, Dongran and Yang, Bo and Liu, Dayou and Wang, Hui and Pan, Shirui},
  journal={Neural Networks},
  year={2023},
  publisher={Elsevier}
}

@article{zhang2021neural,
  title={Neural, symbolic and neural-symbolic reasoning on knowledge graphs},
  author={Zhang, Jing and Chen, Bo and Zhang, Lingxi and Ke, Xirui and Ding, Haipeng},
  journal={AI Open},
  volume={2},
  pages={14--35},
  year={2021},
  publisher={Elsevier}
}

@article{brachman1977s,
  title={What's in a concept: structural foundations for semantic networks},
  author={Brachman, Ronald J},
  journal={International journal of man-machine studies},
  volume={9},
  number={2},
  pages={127--152},
  year={1977},
  publisher={Elsevier}
}

@misc{minsky1974framework,
  title={A framework for representing knowledge},
  author={Minsky, Marvin and others},
  year={1974},
  publisher={Massachusetts Institute of Technology AI Laboratory Cambridge}
}

@article{fillmore2006frame,
  title={Frame semantics},
  author={Fillmore, Charles J and others},
  journal={Cognitive linguistics: Basic readings},
  volume={34},
  pages={373--400},
  year={2006},
  publisher={Mouton de Gruyter Berlin}
}

@article{alam2024neurosymbolic,
  title={Neurosymbolic Methods for Dynamic Knowledge Graphs},
  author={Alam, Mehwish and Gesese, Genet Asefa and Paris, Pierre-Henri},
  journal={arXiv preprint arXiv:2409.04572},
  year={2024}
}

@article{gangemi2020closing,
  title={Closing the Loop between knowledge patterns in cognition and the Semantic Web},
  author={Gangemi, Aldo},
  journal={Semantic Web},
  volume={11},
  number={1},
  pages={139--151},
  year={2020},
  publisher={IOS Press}
}

@book{miller1998wordnet,
  title={WordNet: An electronic lexical database},
  author={Miller, George A},
  year={1998},
  publisher={MIT press}
}

@book{schuler2005verbnet,
  title={VerbNet: A broad-coverage, comprehensive verb lexicon},
  author={Schuler, Karin Kipper},
  year={2005},
  publisher={University of Pennsylvania}
}

@Inproceedings{gangemi2018afaoocm,
  author = "Aldo Gangemi and Valentina Presutti and Mehwish Alam", 
  booktitle = "Formal Ontology in Information Systems - Proceedings of the 10th International Conference, FOIS 2018, Cape Town, South Africa, 19-21 September 2018.", 
  month = "September", 
  publisher = "IOS Press", 
  title = "Amnestic Forgery: An Ontology of Conceptual Metaphors.", 
  year = "2018", 
}

@article{de2022imageschemanet,
  title={Imageschemanet: Formalizing embodied commonsense knowledge providing an imageschematic layer to framester},
  author={De Giorgis, Stefano and Gangemi, Aldo and Gromann, Dagmar},
  journal={Semantic Web Journal, forthcoming},
  year={2022}
}

@inproceedings{navigli2010babelnet,
  title={BabelNet: Building a very large multilingual semantic network},
  author={Navigli, Roberto and Ponzetto, Simone Paolo},
  booktitle={Proceedings of the 48th annual meeting of the association for computational linguistics},
  pages={216--225},
  year={2010}
}

@incollection{auer2007dbpedia,
  title={Dbpedia: A nucleus for a web of open data},
  author={Auer, S{\"o}ren and Bizer, Christian and Kobilarov, Georgi and Lehmann, Jens and Cyganiak, Richard and Ives, Zachary},
  booktitle={The semantic web},
  pages={722--735},
  year={2007},
  publisher={Springer}
}

@inproceedings{suchanek2007yago,
  title={Yago: a core of semantic knowledge},
  author={Suchanek, Fabian M and Kasneci, Gjergji and Weikum, Gerhard},
  booktitle={Proceedings of the 16th international conference on World Wide Web},
  pages={697--706},
  year={2007}
}

@article{gangemi2003sweetening,
  title={Sweetening wordnet with dolce},
  author={Gangemi, Aldo and Guarino, Nicola and Masolo, Claudio and Oltramari, Alessandro},
  journal={AI magazine},
  volume={24},
  number={3},
  pages={13--13},
  year={2003}
}

@article{pan2024unifying,
  title={Unifying large language models and knowledge graphs: A roadmap},
  author={Pan, Shirui and Luo, Linhao and Wang, Yufei and Chen, Chen and Wang, Jiapu and Wu, Xindong},
  journal={IEEE Transactions on Knowledge and Data Engineering},
  year={2024},
  publisher={IEEE}
}

@article{gangemi2017semantic,
  title={Semantic web machine reading with FRED},
  author={Gangemi, Aldo and Presutti, Valentina and Reforgiato Recupero, Diego and Nuzzolese, Andrea Giovanni and Draicchio, Francesco and Mongiov{\`\i}, Misael},
  journal={Semantic Web},
  volume={8},
  number={6},
  pages={873--893},
  year={2017},
  publisher={IOS Press}
}

@inproceedings{baker1998berkeley,
  title={The berkeley framenet project},
  author={Baker, Collin F and Fillmore, Charles J and Lowe, John B},
  booktitle={Proceedings of the 17th international conference on Computational linguistics-Volume 1},
  pages={86--90},
  year={1998},
  organization={Association for Computational Linguistics}
}

@inproceedings{kingsbury2002treebank,
  title={From TreeBank to PropBank.},
  author={Kingsbury, Paul R and Palmer, Martha},
  booktitle={LREC},
  pages={1989--1993},
  year={2002}
}

@inproceedings{karttunen2016presupposition,
  title={Presupposition: What went wrong?},
  author={Karttunen, Lauri},
  booktitle={Semantics and Linguistic Theory},
  pages={705--731},
  year={2016}
}

@misc{frege1892sense,
  title={On sense and reference},
  author={Frege, Gottlob},
  year={1892},
  publisher={na}
}

@article{strawson1950referring,
  title={On referring},
  author={Strawson, Peter F},
  journal={Mind},
  volume={59},
  number={235},
  pages={320--344},
  year={1950},
  publisher={JSTOR}
}

@article{grice1975logic,
  title={Logic and conversation},
  author={Grice, HP},
  journal={Syntax and semantics},
  volume={3},
  year={1975}
}

@book{levinson2000presumptive,
  title={Presumptive meanings: The theory of generalized conversational implicature},
  author={Levinson, Stephen C},
  year={2000},
  publisher={MIT press}
}

@article{kahneman1991article,
  title={Article commentary: Judgment and decision making: A personal view},
  author={Kahneman, Daniel},
  journal={Psychological science},
  volume={2},
  number={3},
  pages={142--145},
  year={1991},
  publisher={SAGE Publications Sage CA: Los Angeles, CA}
}

@book{Johnson87,
	Address = {Chicago and London},
	Author = {Mark Johnson},
	Isbn = {0-226-40318-1},
	Publisher = {The University of Chicago Press},
	Title = {{The Body in the Mind: The Bodily Basis of Meaning, Imagination, and Reason}},
	Year = {1987}
}

@book{lakoff1980metaphors,
  title={Metaphors we live by},
  author={Lakoff, George and Johnson, Mark},
  year={1980},
  publisher={University of Chicago press}
}

@book{lakoff1999philosophy,
  title={Philosophy in the flesh: The embodied mind and its challenge to western thought},
  author={Lakoff, George and Johnson, Mark and others},
  volume={640},
  year={1999},
  publisher={Basic books New York}
}

@article{besold2017narrative,
  title={A narrative in three acts: Using combinations of image schemas to model events},
  author={Besold, Tarek R and Hedblom, Maria M and Kutz, Oliver},
  journal={Biologically inspired cognitive architectures},
  volume={19},
  pages={10--20},
  year={2017},
  publisher={Elsevier}
}

@book{hedblom2020image,
  title={Image schemas and concept invention: cognitive, logical, and linguistic investigations},
  author={Hedblom, Maria M},
  year={2020},
  publisher={Springer Nature}
}

@article{maudslay2024chainnet,
  title={ChainNet: Structured Metaphor and Metonymy in WordNet},
  author={Maudslay, Rowan Hall and Teufel, Simone and Bond, Francis and Pustejovsky, James},
  journal={arXiv preprint arXiv:2403.20308},
  year={2024}
}

@incollection{graham2013moral,
  title={Moral foundations theory: The pragmatic validity of moral pluralism},
  author={Graham, Jesse and Haidt, Jonathan and Koleva, Sena and Motyl, Matt and Iyer, Ravi and Wojcik, Sean P and Ditto, Peter H},
  booktitle={Advances in experimental social psychology},
  volume={47},
  pages={55--130},
  year={2013},
  publisher={Elsevier}
}

@article{rozin_cad_1999,
	title = {The {CAD} triad hypothesis: a mapping between three moral emotions (contempt, anger, disgust) and three moral codes (community, autonomy, divinity).},
	volume = {76},
	number = {4},
	journal = {Journal of personality and social psychology},
	author = {Rozin, Paul and Lowery, Laura and Imada, Sumio and Haidt, Jonathan},
	year = {1999},
	note = {Publisher: American Psychological Association},
	pages = {574},
}

@article{schwartz_extending_2001,
	title = {Extending the cross-cultural validity of the theory of basic human values with a different method of measurement},
	volume = {32},
	number = {5},
	journal = {Journal of cross-cultural psychology},
	author = {Schwartz, Shalom H and Melech, Gila and Lehmann, Arielle and Burgess, Steven and Harris, Mari and Owens, Vicki},
	year = {2001},
	note = {Publisher: Sage Publications Sage CA: Thousand Oaks, CA},
	pages = {519--542},
}

@article{peirce1902logic,
  title={Logic as semiotic: The theory of signs},
  author={Peirce, Charles Sanders and Buchler, Justus},
  journal={Philosophical Writings of Peirce, ed. Justus Buchler (New York: Dover, 1955)},
  pages={100},
  year={1902}
}

@article{eliot1920hamlet,
  title={Hamlet and his problems},
  author={Eliot, Thomas Stearns and others},
  journal={The sacred wood: Essays on poetry and criticism},
  volume={4},
  pages={95--104},
  year={1920},
  publisher={Methuen London}
}

@article{forrester1971counterintuitive,
  title={Counterintuitive behavior of social systems},
  author={Forrester, Jay W},
  journal={Theory and decision},
  volume={2},
  number={2},
  pages={109--140},
  year={1971},
  publisher={Springer}
}

@inproceedings{gangemi2023taf,
  title={{Text2AMR2FRED, a Tool for Transforming Text into RDF/OWL Knowledge Graphs via Abstract Meaning Representation}},
  author={Gangemi, Aldo and Graciotti, Arianna and Meloni, Antonello and Nuzzolese, Andrea Giovanni and Presutti, Valentina and Reforgiato Recupero, Diego and Russo, Alessandro and Tripodi, Rocco},
  booktitle={Proceedings of the ISWC 2023 Posters, Demos and Industry Tracks: From Novel Ideas to Industrial Practice
co-located with 22nd International Semantic Web Conference (ISWC 2023)},
  year={2023}
}

@article{ha2018world,
  title={World models},
  author={Ha, David and Schmidhuber, J{\"u}rgen},
  journal={arXiv preprint arXiv:1803.10122},
  year={2018}
}

@article{nolfi2023unexpected,
  title={On the unexpected abilities of large language models},
  author={Nolfi, Stefano},
  journal={Adaptive Behavior},
  pages={10597123241256754},
  year={2023},
  publisher={SAGE Publications Sage UK: London, England}
}

@incollection{lambek1988categorial,
  title={Categorial and categorical grammars},
  author={Lambek, Joachim},
  booktitle={Categorial grammars and natural language structures},
  pages={297--317},
  year={1988},
  publisher={Springer}
}

@inproceedings{bos2008wide,
  title={Wide-coverage semantic analysis with boxer},
  author={Bos, Johan},
  booktitle={Semantics in text processing. step 2008 conference proceedings},
  pages={277--286},
  year={2008}
}

@article{grau2008owl,
  title={OWL 2: The next step for OWL},
  author={Grau, Bernardo Cuenca and Horrocks, Ian and Motik, Boris and Parsia, Bijan and Patel-Schneider, Peter and Sattler, Ulrike},
  journal={Journal of Web Semantics},
  volume={6},
  number={4},
  pages={309--322},
  year={2008},
  publisher={Elsevier}
}

@inproceedings{de2022basic,
  title={Basic human values and moral foundations theory in valuenet ontology},
  author={De Giorgis, Stefano and Gangemi, Aldo and Damiano, Rossana},
  booktitle={International conference on knowledge engineering and knowledge management},
  pages={3--18},
  year={2022},
  organization={Springer}
}

@inproceedings{bast2017qlever,
  title={Qlever: A query engine for efficient sparql+ text search},
  author={Bast, Hannah and Buchhold, Bj{\"o}rn},
  booktitle={Proceedings of the 2017 ACM on Conference on Information and Knowledge Management},
  pages={647--656},
  year={2017}
}



\begin{thebibliography}{99}

\bibitem{meyer2023llm}
Meyer, L.P., Stadler, C., Frey, J., Radtke, N., Junghanns, K., Meissner, R., Dziwis, G., Bulert, K., Martin, M.,
``LLM-assisted knowledge graph engineering: Experiments with ChatGPT,''
in Working conference on Artificial Intelligence Development for a Resilient and Sustainable Tomorrow,
pp. 103--115, 2023.

\bibitem{gangemi2016kbs}
Gangemi, A., Recupero, D.R., Mongiov{\`\i}, M., Nuzzolese, A.G., Presutti, V.,
``Identifying motifs for evaluating open knowledge extraction on the Web,''
Knowledge Based Systems,
vol. 108, pp. 33--41, 2016.

\bibitem{zhu2024llms}
Zhu, Y., Wang, X., Chen, J., Qiao, S., Ou, Y., Yao, Y., Deng, S., Chen, H., Zhang, N.,
``LLMs for knowledge graph construction and reasoning: Recent capabilities and future opportunities,''
World Wide Web,
vol. 27, no. 5, pp. 58, 2024.

\bibitem{sen2023knowledge}
Sen, P., Mavadia, S., Saffari, A.,
``Knowledge graph-augmented language models for complex question answering,''
in Proceedings of the 1st Workshop on Natural Language Reasoning and Structured Explanations (NLRSE),
pp. 1--8, 2023.

\bibitem{edge2024local}
Edge, D., Trinh, H., Cheng, N., Bradley, J., Chao, A., Mody, A., Truitt, S., Larson, J.,
``From local to global: A graph rag approach to query-focused summarization,''
arXiv preprint arXiv:2404.16130,
2024.

\bibitem{yao2019kg}
Yao, L., Mao, C., Luo, Y.,
``KG-BERT: BERT for knowledge graph completion,''
arXiv preprint arXiv:1909.03193,
2019.

\bibitem{lu2020utilizing}
Lu, F., Cong, P., Huang, X.,
``Utilizing textual information in knowledge graph embedding: A survey of methods and applications,''
IEEE Access,
vol. 8, pp. 92072--92088, 2020.

\bibitem{tiwari2022preface}
Tiwari, S., Mihindukulasooriya, N., Osborne, F., Kontokostas, D., D'Souza, J., Kejriwal, M., et al.,
``Preface for the International Workshop on Knowledge Graph Generation from Text,''
in CEUR WORKSHOP PROCEEDINGS,
vol. 3184, 2022.

\bibitem{poveda2014oops}
Poveda-Villal{\'o}n, M., G{\'o}mez-P{\'e}rez, A., Su{\'a}rez-Figueroa, M.C.,
``Oops!(ontology pitfall scanner!): An on-line tool for ontology evaluation,''
International Journal on Semantic Web and Information Systems (IJSWIS),
vol. 10, no. 2, pp. 7--34, 2014.

\bibitem{fathallah2024neon}
Fathallah, N., Das, A., De Giorgis, S., Poltronieri, A., Haase, P., Kovriguina, L.,
``NeOn-GPT: A Large Language Model-Powered Pipeline for Ontology Learning,''
in The Extended Semantic Web Conference,
2024.

\bibitem{glimm2014hermit}
Glimm, B., Horrocks, I., Motik, B., Stoilos, G., Wang, Z.,
``HermiT: an OWL 2 reasoner,''
Journal of automated reasoning,
vol. 53, pp. 245--269, 2014.

\bibitem{paulheim2015serving}
Paulheim, H., Gangemi, A.,
``Serving DBpedia with DOLCE--more than just adding a cherry on top,''
in The Semantic Web-ISWC 2015: 14th International Semantic Web Conference,
pp. 180--196, 2015.

\bibitem{bevilacqua2021one}
Bevilacqua, M., Blloshmi, R., Navigli, R.,
``One SPRING to rule them both: Symmetric AMR semantic parsing and generation without a complex pipeline,''
in Proceedings of the AAAI Conference on Artificial Intelligence,
vol. 35, no. 14, pp. 12564--12573, 2021.

\bibitem{wu2019scalable}
Wu, L., Petroni, F., Josifoski, M., Riedel, S., Zettlemoyer, L.,
``Scalable zero-shot entity linking with dense entity retrieval,''
arXiv preprint arXiv:1911.03814,
2019.

\bibitem{meloni2017amr2fred}
Meloni, A., Reforgiato Recupero, D., Gangemi, A.,
``AMR2FRED, a tool for translating abstract meaning representation to motif-based linguistic knowledge graphs,''
in The Semantic Web: ESWC 2017 Satellite Events,
pp. 43--47, 2017.

\bibitem{bevilacqua-navigli-2020-breaking}
Bevilacqua, M., Navigli, R.,
``Breaking Through the 80\% Glass Ceiling: Raising the State of the Art in Word Sense Disambiguation by Incorporating Knowledge Graph Information,''
in Proceedings of the 58th Annual Meeting of the Association for Computational Linguistics,
pp. 2854--2864, 2020.

\bibitem{hamilton2022neuro}
Hamilton, K., Nayak, A., Bo{\v{z}}i{\'c}, B., Longo, L.,
``Is neuro-symbolic ai meeting its promises in natural language processing? a structured review,''
Semantic Web,
pp. 1--42, 2022.

\bibitem{kautz2022third}
Kautz, H.,
``The third ai summer: AAAI robert s. engelmore memorial lecture,''
AI magazine,
vol. 43, no. 1, pp. 105--125, 2022.

\bibitem{gangemi2016framester}
Gangemi, A., Alam, M., Asprino, L., Presutti, V., Recupero, D.R.,
``Framester: A wide coverage linguistic linked data hub,''
in Knowledge Engineering and Knowledge Management,
pp. 239--254, 2016.

\bibitem{belle2020symbolic}
Belle, V.,
``Symbolic logic meets machine learning: A brief survey in infinite domains,''
in International conference on scalable uncertainty management,
pp. 3--16, 2020.

\bibitem{besold2021neural}
Besold, T.R., d'Avila Garcez, A., Bader, S., Bowman, H., Domingos, P., Hitzler, P., et al.,
``Neural-symbolic learning and reasoning: A survey and interpretation 1,''
in Neuro-Symbolic Artificial Intelligence: The State of the Art,
pp. 1--51, 2021.

\bibitem{garcez2023neurosymbolic}
Garcez, A.d'A., Lamb, L.C.,
``Neurosymbolic AI: The 3rd wave,''
Artificial Intelligence Review,
vol. 56, no. 11, pp. 12387--12406, 2023.

\bibitem{lamb2020graph}
Lamb, L.C., Garcez, A., Gori, M., Prates, M., Avelar, P., Vardi, M.,
``Graph neural networks meet neural-symbolic computing: A survey and perspective,''
arXiv preprint arXiv:2003.00330,
2020.

\bibitem{sarker2021neuro}
Sarker, M.K., Zhou, L., Eberhart, A., Hitzler, P.,
``Neuro-symbolic artificial intelligence,''
AI Communications,
vol. 34, no. 3, pp. 197--209, 2021.

\bibitem{yao2018learning}
Yao, Y., Xu, J., Shi, J., Xu, B.,
``Learning to activate logic rules for textual reasoning,''
Neural Networks,
vol. 106, pp. 42--49, 2018.

\bibitem{yu2023survey}
Yu, D., Yang, B., Liu, D., Wang, H., Pan, S.,
``A survey on neural-symbolic learning systems,''
Neural Networks,
2023.

\bibitem{zhang2021neural}
Zhang, J., Chen, B., Zhang, L., Ke, X., Ding, H.,
``Neural, symbolic and neural-symbolic reasoning on knowledge graphs,''
AI Open,
vol. 2, pp. 14--35, 2021.

\bibitem{brachman1977s}
Brachman, R.J.,
``What's in a concept: structural foundations for semantic networks,''
International journal of man-machine studies,
vol. 9, no. 2, pp. 127--152, 1977.

\bibitem{minsky1974framework}
Minsky, M., et al.,
``A framework for representing knowledge,''
Massachusetts Institute of Technology AI Laboratory Cambridge,
1974.

\bibitem{fillmore2006frame}
Fillmore, C.J., et al.,
``Frame semantics,''
Cognitive linguistics: Basic readings,
vol. 34, pp. 373--400, 2006.

\bibitem{alam2024neurosymbolic}
Alam, M., Gesese, G.A., Paris, P.H.,
``Neurosymbolic Methods for Dynamic Knowledge Graphs,''
arXiv preprint arXiv:2409.04572,
2024.

\bibitem{gangemi2020closing}
Gangemi, A.,
``Closing the Loop between knowledge patterns in cognition and the Semantic Web,''
Semantic Web,
vol. 11, no. 1, pp. 139--151, 2020.

\bibitem{miller1998wordnet}
Miller, G.A.,
``WordNet: An electronic lexical database,''
MIT press,
1998.

\bibitem{schuler2005verbnet}
Schuler, K.K.,
``VerbNet: A broad-coverage, comprehensive verb lexicon,''
University of Pennsylvania,
2005.

\bibitem{gangemi2018afaoocm}
Gangemi, A., Presutti, V., Alam, M.,
``Amnestic Forgery: An Ontology of Conceptual Metaphors,''
in Formal Ontology in Information Systems,
2018.

\bibitem{de2022imageschemanet}
De Giorgis, S., Gangemi, A., Gromann, D.,
``Imageschemanet: Formalizing embodied commonsense knowledge providing an imageschematic layer to framester,''
Semantic Web Journal,
2022.

\bibitem{navigli2010babelnet}
Navigli, R., Ponzetto, S.P.,
``BabelNet: Building a very large multilingual semantic network,''
in Proceedings of the 48th annual meeting of the association for computational linguistics,
pp. 216--225, 2010.

\bibitem{auer2007dbpedia}
Auer, S., Bizer, C., Kobilarov, G., Lehmann, J., Cyganiak, R., Ives, Z.,
``Dbpedia: A nucleus for a web of open data,''
in The semantic web,
pp. 722--735, 2007.

\bibitem{suchanek2007yago}
Suchanek, F.M., Kasneci, G., Weikum, G.,
``Yago: a core of semantic knowledge,''
in Proceedings of the 16th international conference on World Wide Web,
pp. 697--706, 2007.

\bibitem{gangemi2003sweetening}
Gangemi, A., Guarino, N., Masolo, C., Oltramari, A.,
``Sweetening wordnet with dolce,''
AI magazine,
vol. 24, no. 3, pp. 13--13, 2003.

\bibitem{pan2024unifying}
Pan, S., Luo, L., Wang, Y., Chen, C., Wang, J., Wu, X.,
``Unifying large language models and knowledge graphs: A roadmap,''
IEEE Transactions on Knowledge and Data Engineering,
2024.

\bibitem{gangemi2017semantic}
Gangemi, A., Presutti, V., Reforgiato Recupero, D., Nuzzolese, A.G., Draicchio, F., Mongiov{\`\i}, M.,
``Semantic web machine reading with FRED,''
Semantic Web,
vol. 8, no. 6, pp. 873--893, 2017.

\bibitem{baker1998berkeley}
Baker, C.F., Fillmore, C.J., Lowe, J.B.,
``The berkeley framenet project,''
in Proceedings of the 17th international conference on Computational linguistics,
pp. 86--90, 1998.

\bibitem{kingsbury2002treebank}
Kingsbury, P.R., Palmer, M.,
``From TreeBank to PropBank,''
in LREC,
pp. 1989--1993, 2002.

\bibitem{karttunen2016presupposition}
Karttunen, L.,
``Presupposition: What went wrong?,''
in Semantics and Linguistic Theory,
pp. 705--731, 2016.

\bibitem{frege1892sense}
Frege, G.,
``On sense and reference,''
1892.

\bibitem{strawson1950referring}
Strawson, P.F.,
``On referring,''
Mind,
vol. 59, no. 235, pp. 320--344, 1950.

\bibitem{grice1975logic}
Grice, H.P.,
``Logic and conversation,''
Syntax and semantics,
vol. 3, 1975.

\bibitem{levinson2000presumptive}
Levinson, S.C.,
``Presumptive meanings: The theory of generalized conversational implicature,''
MIT press,
2000.

\bibitem{kahneman1991article}
Kahneman, D.,
``Article commentary: Judgment and decision making: A personal view,''
Psychological science,
vol. 2, no. 3, pp. 142--145, 1991.

\bibitem{Johnson87}
Johnson, M.,
``The Body in the Mind: The Bodily Basis of Meaning, Imagination, and Reason,''
The University of Chicago Press,
1987.

\bibitem{lakoff1980metaphors}
Lakoff, G., Johnson, M.,
``Metaphors we live by,''
University of Chicago press,
1980.

\bibitem{lakoff1999philosophy}
Lakoff, G., Johnson, M., et al.,
``Philosophy in the flesh: The embodied mind and its challenge to western thought,''
Basic books New York,
1999.

\bibitem{besold2017narrative}
Besold, T.R., Hedblom, M.M., Kutz, O.,
``A narrative in three acts: Using combinations of image schemas to model events,''
Biologically inspired cognitive architectures,
vol. 19, pp. 10--20, 2017.

\bibitem{hedblom2020image}
Hedblom, M.M.,
``Image schemas and concept invention: cognitive, logical, and linguistic investigations,''
Springer Nature,
2020.

\bibitem{maudslay2024chainnet}
Maudslay, R.H., Teufel, S., Bond, F., Pustejovsky, J.,
``ChainNet: Structured Metaphor and Metonymy in WordNet,''
arXiv preprint arXiv:2403.20308,
2024.

\bibitem{graham2013moral}
Graham, J., Haidt, J., Koleva, S., Motyl, M., Iyer, R., Wojcik, S.P., Ditto, P.H.,
``Moral foundations theory: The pragmatic validity of moral pluralism,''
Advances in experimental social psychology,
vol. 47, pp. 55--130, 2013.

\bibitem{rozin_cad_1999}
Rozin, P., Lowery, L., Imada, S., Haidt, J.,
``The CAD triad hypothesis: a mapping between three moral emotions (contempt, anger, disgust) and three moral codes (community, autonomy, divinity),''
Journal of personality and social psychology,
vol. 76, no. 4, pp. 574, 1999.

\bibitem{schwartz_extending_2001}
Schwartz, S.H., Melech, G., Lehmann, A., Burgess, S., Harris, M., Owens, V.,
``Extending the cross-cultural validity of the theory of basic human values with a different method of measurement,''
Journal of cross-cultural psychology,
vol. 32, no. 5, pp. 519--542, 2001.

\bibitem{peirce1902logic}
Peirce, C.S., Buchler, J.,
``Logic as semiotic: The theory of signs,''
Philosophical Writings of Peirce,
pp. 100, 1902.

\bibitem{eliot1920hamlet}
Eliot, T.S., et al.,
``Hamlet and his problems,''
The sacred wood: Essays on poetry and criticism,
vol. 4, pp. 95--104, 1920.

\bibitem{forrester1971counterintuitive}
Forrester, J.W.,
``Counterintuitive behavior of social systems,''
Theory and decision,
vol. 2, no. 2, pp. 109--140, 1971.

\bibitem{gangemi2023taf}
Gangemi, A., Graciotti, A., Meloni, A., Nuzzolese, A.G., Presutti, V., Reforgiato Recupero, D., Russo, A., Tripodi, R.,
``Text2AMR2FRED, a Tool for Transforming Text into RDF/OWL Knowledge Graphs via Abstract Meaning Representation,''
in Proceedings of the ISWC 2023 Posters, Demos and Industry Tracks,
2023.

\bibitem{ha2018world}
Ha, D., Schmidhuber, J.,
``World models,''
arXiv preprint arXiv:1803.10122,
2018.

\bibitem{nolfi2023unexpected}
Nolfi, S.,
``On the unexpected abilities of large language models,''
Adaptive Behavior,
2023.

\bibitem{lambek1988categorial}
Lambek, J.,
``Categorial and categorical grammars,''
in Categorial grammars and natural language structures,
pp. 297--317, 1988.

\bibitem{bos2008wide}
Bos, J.,
``Wide-coverage semantic analysis with boxer,''
in Semantics in text processing. STEP 2008 conference proceedings,
pp. 277--286, 2008.

\bibitem{grau2008owl}
Grau, B.C., Horrocks, I., Motik, B., Parsia, B., Patel-Schneider, P., Sattler, U.,
``OWL 2: The next step for OWL,''
Journal of Web Semantics,
vol. 6, no. 4, pp. 309--322, 2008.

\bibitem{de2022basic}
De Giorgis, S., Gangemi, A., Damiano, R.,
``Basic human values and moral foundations theory in valuenet ontology,''
in International conference on knowledge engineering and knowledge management,
pp. 3--18, 2022.

\bibitem{bast2017qlever}
Bast, H., Buchhold, B.,
``Qlever: A query engine for efficient sparql+ text search,''
in Proceedings of the 2017 ACM on Conference on Information and Knowledge Management,
pp. 647--656, 2017.

\end{thebibliography}
